%% file: main.tex
\documentclass[twocolumn]{article}

\usepackage{arxiv}

\usepackage[utf8]{inputenc} 
\usepackage[T1]{fontenc}    
\usepackage{hyperref}       
\usepackage{url}            
\usepackage{booktabs}       
\usepackage{amsmath,amsfonts}       
\usepackage{nicefrac}       
\usepackage{microtype}      
\usepackage{cleveref}       
\usepackage{lipsum}         
\usepackage{graphicx}
\usepackage{doi}
\usepackage{adjustbox}
\usepackage[permil]{overpic}
\usepackage{multirow}
\usepackage{xcolor}

\title{First Shape, Then Meaning: Efficient Geometry and Semantics Learning for Indoor Reconstruction}

\usepackage{authblk}

\author[1,2]{%
	Remi Chierchia\thanks{\texttt{remi.chierchia@hdr.qut.edu.au}}%
}
\author[1]{%
	L\'eo Lebrat%
}
\author[1,2]{%
	David Ahmedt-Aristizabal%
}
\author[1]{%
	Olivier Salvado%
}

\author[1]{%
	\\Clinton Fookes%
}
\author[1]{%
	Rodrigo Santa Cruz%
}

\affil[1]{\small School of Electrical Engineering \& Robotics, Queensland University of Technology, Australia}
\affil[2]{\small Imaging and Computer Vision Group, CSIRO Data61, Australia}

\newcommand{\NAME}{\textit{FSTM}}

\begin{document}
\maketitle

\begin{abstract}
Neural Surface Reconstruction has become a standard methodology for indoor 3D reconstruction, with Signed Distance Functions (SDFs) proving particularly effective for representing scene geometry. 
A variety of applications require a detailed understanding of the scene context, driving the need for object-level semantic signals.
While recent methods successfully integrate semantic labels, they often inherit the slow training time and limited scalability of multi-SDF learning.
In this paper, we introduce \NAME{}, a unified approach for learning geometry and semantics through a two-step process: a geometry warm-up using RGB inputs and geometric cues, followed by semantic field estimation.
By first optimising geometry without semantic supervision, we observe substantial improvements compared to the standard joint optimisation.
Rather than relying on specialised modules or complex multi-SDF designs, \NAME{} shows that a streamlined formulation is sufficient to achieve strong geometric and semantic reconstructions.
Experiments on both synthetic and real-world indoor datasets show that our method outperforms multi-SDF approaches. It trains 2.3$\times$ faster on Replica, improves robustness to real-world imperfections on ScanNet++, and achieves higher recall by recovering the surfaces of more objects in the scene.
The code will be made available at~\href{https://remichierchia.github.io/FSTM/}{remichierchia.github.io/FSTM}.
\end{abstract}

\vspace{-2em}

\section{Introduction}
\label{sec:intro}




Accurate 3D reconstruction is essential for applications ranging from robotics ~\cite{rosinol2023nerf,adamkiewicz2022vision} to  digital asset creation~\cite{lin2023magic3d} and healthcare~\cite{zha2023endosurf,chierchia_2025_wacv}. 
Recent advances in neural surface reconstruction based on Signed Distance Function (SDF) learning~\cite{wang2021neus,NEURIPS2020_1a77befc} have enabled the recovery of 3D geometry from multi-view images.
While integrating geometric priors such as depth and normal maps further refines 3D reconstructions~\cite{Yu2022SDFStudio}, purely geometric representations fall short for downstream tasks that require scene understanding, object interation, and editing.

Despite the extensive study of geometric cues, the role of semantic priors in neural surface reconstruction remains relatively underexplored.
Semantics do not merely label a scene, they provide powerful regularisation, encoding object-level structural details that actively enhance reconstruction quality and robustness against real-world noise.
This motivates our goal of designing a unified neural surface reconstruction framework to jointly estimate accurate geometry and per-object semantic labels, tackling the crucial challenge of building semantically rich 3D environments.

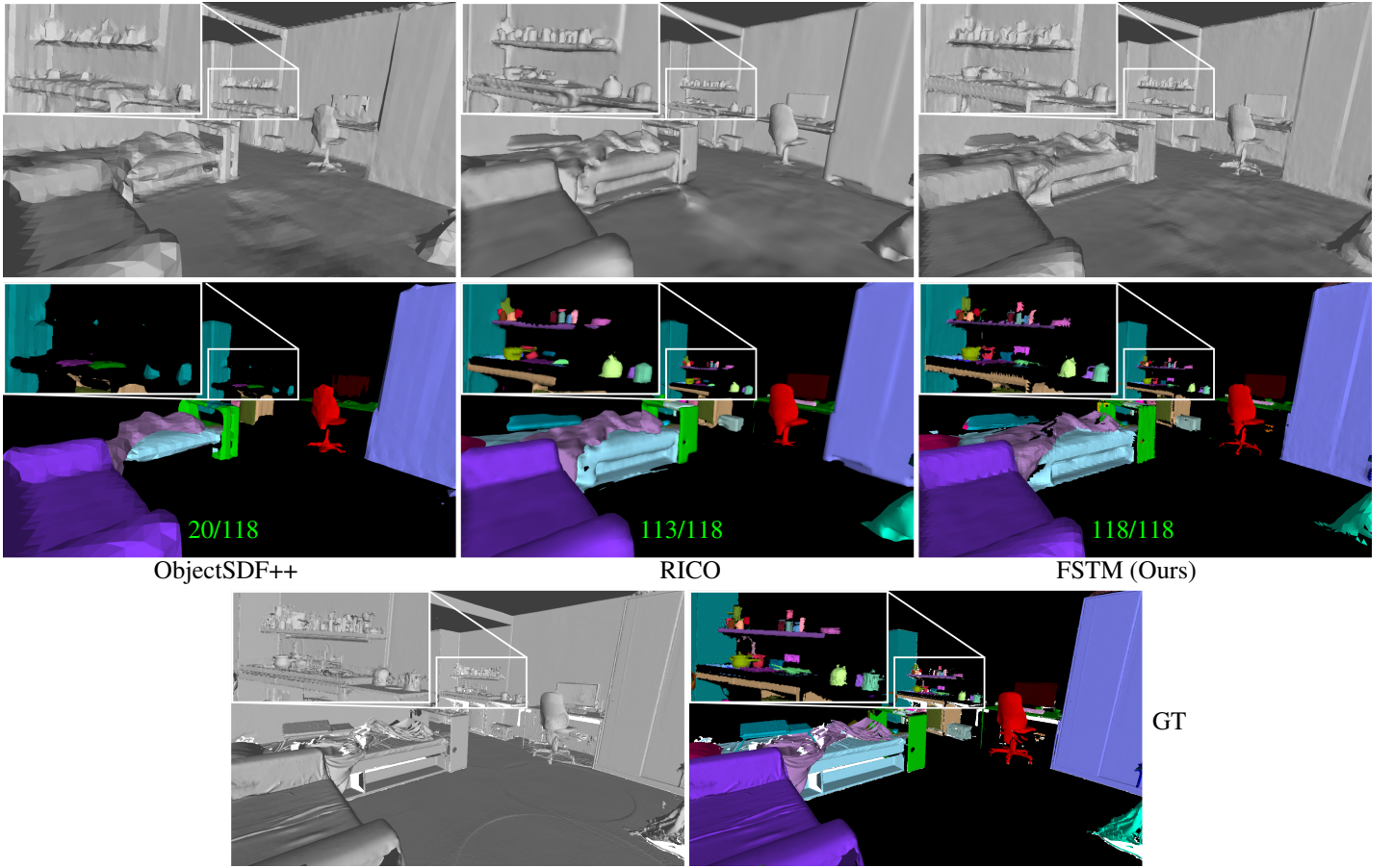
\begin{figure*}[!t]

\begin{overpic}[width=\textwidth,percent]{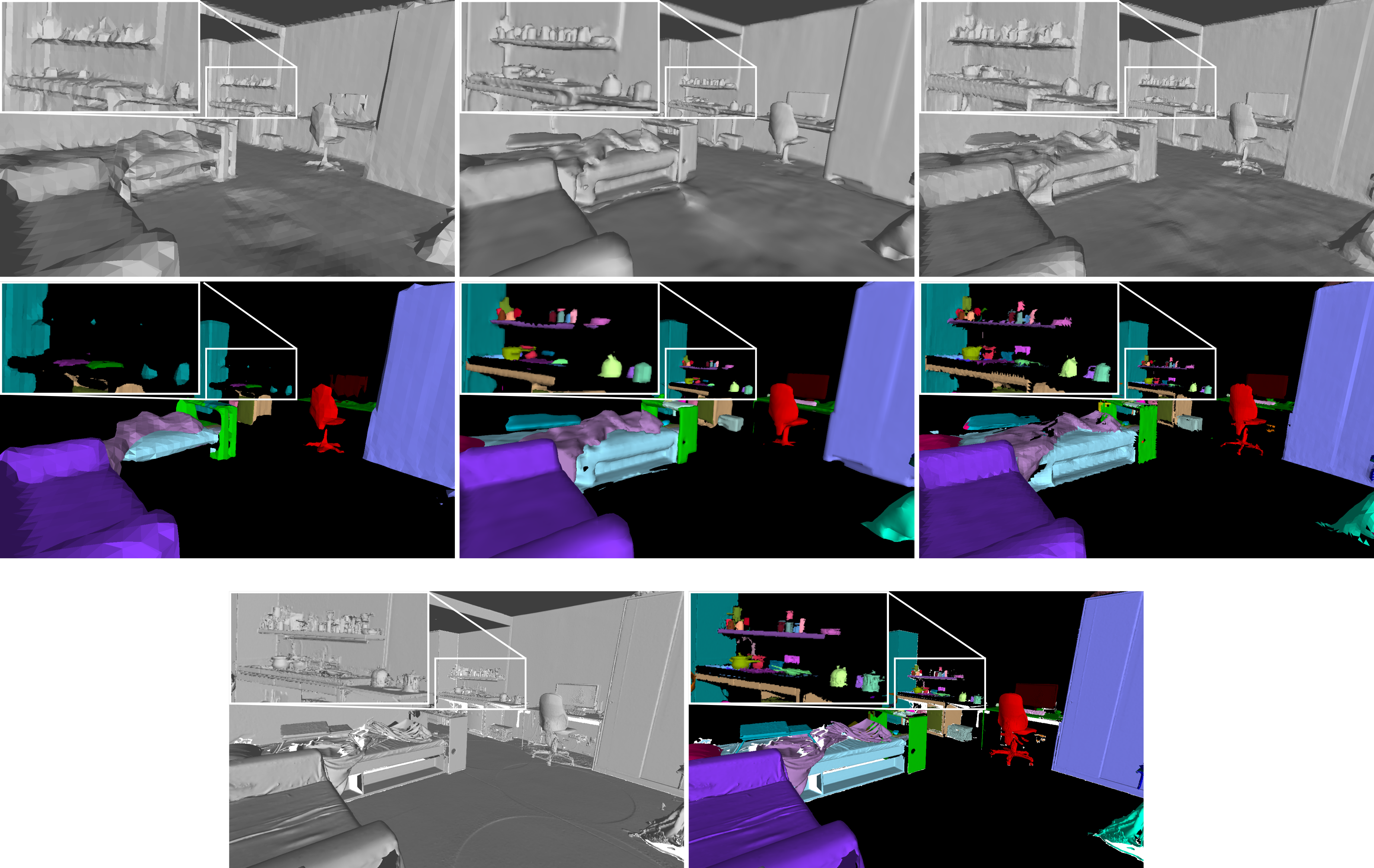}

    \put (13.5,24) {\normalsize \textcolor{green}{20/118}}
    \put (46.5,24) {\normalsize \textcolor{green}{113/118}}
    \put (79.5,24) {\normalsize \textcolor{green}{118/118}}
    
    \put (48,21) {RICO}
    \put (11,21) {\normalsize ObjectSDF++}
    \put (77,21) {\normalsize FSTM (Ours)}
    \put (84,10) {\normalsize GT}
    
\end{overpic}
\caption{Qualitative comparison of 3D object segmentation for the ScanNet++ dataset of scene \textit{8be0cd3817}. FSTM allows for a more robust reconstruction, particularly in recovering small objects. In green text, the number of objects retrieved by each method out of the total number of objects in the scene.}
\label{fig:scannetpp_comparison}

\end{figure*}

Approaches like ObjectSDF++~\cite{Wu_2023_ICCV} and RICO~\cite{Li_2023_ICCV} address this by decomposing scenes into individual object instances, representing each with a separate SDF. 
The reliance on individual SDFs inherently leads to high computational overhead and optimisation instability during training. 
Furthermore, although recent methods like RICO successfully mitigate some of the degradation caused by occlusions, multi-SDF architectures still struggle to maintain consistent detail and recall as overall scene complexity scales. Consequently, they often underperform in highly cluttered real-world settings, missing instances or generating oversimplified geometry, as illustrated in Figure~\ref{fig:scannetpp_comparison}.

These challenges underscore the necessity for a more efficient approach to incorporating semantics into neural surface learning.
We propose \NAME{}, a method that maintains generality by estimating a single SDF for the entire scene, optimising both geometry and semantics.
However, early joint optimisation of the semantic and geometric fields negatively impacts reconstruction quality.
To address this challenge, we present a two-step training strategy comprising: (i) a geometry warm-up phase supervised by RGB images and monocular priors, followed by (ii) joint training of both geometry and semantic networks.
Finally, we propose a lightweight 3D segmentation technique to directly query the learned semantic field, enabling dense, object-level labelling of the reconstructed mesh directly at test time.

We evaluate \NAME{} on two widely used indoor 3D reconstruction datasets, Replica and ScanNet++. 
Our method outperforms ObjectSDF++ and RICO in both scene- and object-level reconstruction tasks, while training 2.3$\times$ faster on Replica.
%
%
Additionally, \NAME{} demonstrates enhanced robustness in complex, multi-object scenes by reconstructing up to six times more objects than prior methods. 
We demonstrate that integrating semantic supervision via our two-step optimisation significantly improves reconstruction accuracy over MonoSDF under the same training budget, highlighting the crucial role of semantics in neural surface reconstruction.

\section{Related Work}
\label{sec:related_work}

We first review neural surface reconstruction methods foundational to modern 3D reconstruction~\cite{xiao2025neural}. 
Next, we examine approaches that integrate semantic understanding into neural implicit representations, highlighting the design limitations of multi-SDF formulations. 
Lastly, we survey recent semantic extensions of 3D Gaussian splatting, situating our method within the broader landscape of object-aware scene modelling before motivating our design.

Implicit neural representations, particularly signed distance functions (SDFs), have shown superior accuracy over traditional mesh or point cloud representations~\cite{wang2021neus,Li_2023_CVPR}. To improve reconstruction quality, cues such as SfM and multi-view patch correspondences~\cite{NEURIPS2022_16415eed,Zhang_2021_ICCV,Darmon_2022_CVPR}, surface smoothness~\cite{Zhang_2022_CVPR,Niemeyer_2022_CVPR,Oechsle_2021_ICCV}, and monocular priors~\cite{NEURIPS2022_9f0b1220,wang2022neuris} are commonly integrated. Our approach extends this line by incorporating semantic information for enhanced 3D scene understanding. While semantics are applied in other contexts such as SLAM and 3D perception~\cite{zhu2024sni,liu2024surroundsdf}, these are beyond our scope, which focuses on neural surface reconstruction for object-aware scene modelling.

Integrating semantic understanding into 3D representations enables downstream applications such as scene editing and understanding~\cite{ov_nerf2024,kobayashi2022decomposing,NEURIPS2023_525d2440,Xie_2023_ICCV,repaint_nerf,Haque_2023_ICCV}.
Early methods extended NeRF architectures to jointly model appearance and semantics, with Semantic-NeRF~\cite{Zhi_2021_ICCV} recovering consistent 2D segmentation from noisy labels. Subsequent works advanced 3D panoptic segmentation using coarse annotations~\cite{panoptic_nerf,Siddiqui_2023_CVPR,wu2024clusteringsdf}, while others leverage semantics to improve reconstruction under challenging conditions, such as limited views or inaccurate poses~\cite{Jain_2021_ICCV,xu2022sinnerf,wu2024clusteringsdf}. 

In the context of scene generalisation, NeSF~\cite{nesf} learns a mapping function from NeRF representations into generic features, which deep learning models can then segment. S-Ray~\cite{Liu_2023_CVPR} employs cross-reprojection attention modules to strengthen generalisation and robustness in real-world datasets.
GNeSF removes the need for pretrained scene-specific NeRFs, enabling direct inference of 3D segmentations on novel scenes without optimisation. 
These methods perform scene segmentation, prioritising accuracy over reconstruction quality. 
In contrast, we emphasise leveraging semantic priors to enhance geometric fidelity in NeRF-based SDF reconstruction of both scenes and their objects.

Closely related to our work, Manhattan-SDF~\cite{Guo_2022_CVPR} leverages segmentation maps to improve the reconstruction of planar regions common in indoor environments, by enforcing wall orthogonality and alignment with the default coordinate system.
In~\cite{park2024h2o}, the authors introduce an adaptive weighting scheme between normal and colour supervision based on surface texture characteristics, employing a two-phase pipeline with a single object-background mask.
Inspired by object-compositional approaches, ObjectSDF~\cite{wu2022object} and ObjectSDF++~\cite{Wu_2023_ICCV} introduce object-driven losses in a multi-SDF formulation to achieve high object reconstruction accuracy. RICO~\cite{Li_2023_ICCV} introduces a background-to-object constraint to regularise surfaces in partially observed regions. 
Recent work~\cite{Lyu_2024_CVPR} incorporates vision transformers and interactive refinement to improve scene decomposition.
Furthermore, ~\cite{NEURIPS2024_2d880acd} utilises physics simulation to supervise reconstructions, while ~\cite{Ni_2025_CVPR} leverages generative priors to improve geometric fidelity.

In parallel, 3D Gaussian splatting~\cite{kerbl20233d} has emerged as an efficient alternative to neural fields for scene reconstruction and rendering, and has rapidly become a common baseline for semantic and panoptic 3D scene modelling. Gaussian Grouping~\cite{ye2024gaussian}, PLGS~\cite{wang2025plgs}, and PanopticSplatting~\cite{xie2025panopticsplatting} extend 3DGS with object grouping, panoptic lifting, and end-to-end segmentation mechanisms, respectively, primarily targeting robust scene understanding under noisy 2D supervision. Most closely related to our setting, ObjectGS~\cite{zhu2025objectgs} introduces object-aware anchors that adapt Gaussian semantics during reconstruction, explicitly coupling object-aware scene reconstruction with scene understanding. However, Gaussian-based methods generally prioritise editability or open-vocabulary capability over accurate surface modelling.

In contrast to Gaussian-based methods that often sacrifice reconstruction accuracy, \NAME{} is designed specifically to maintain high-fidelity surface extraction. Furthermore, to address the computational complexity and missed detection of small objects inherent in multi-SDF formulations, our approach maintains a single SDF for the entire scene. By preserving object-level accuracy through segmentation priors, without relying on complex multi-object coordination, this streamlined design demonstrates that high-quality, object-aware reconstruction can be achieved efficiently, even in challenging real-world environments.

\section{Limitations of Multi-SDF Designs} 

A standard way to decompose scenes into individual objects is to use instance mask guidance~\cite{Wu_2023_ICCV,wu2022object,Yang_2021_ICCV}.
For example, ObjectSDF++ employs a shallow MLP that outputs an $N$-dimensional vector, where each dimension encodes an object-level SDF.
The whole-scene SDF, required to optimise NeRF’s objective, is then obtained via $\min$ pooling over the predicted object SDFs:
\begin{equation}
    \text{SDF}_{\theta}(\textbf{x}) = \min_{i=1,...,N} \text{SDF}^i_{\theta}(\textbf{x})\,.
\end{equation}
This learnable function parametrised by $\theta$ is trained such that its zero-level set $\Omega = \{ \textbf{x}\in \mathbb{R}^3\  | \  \text{SDF}_{\theta}(x) = 0\}$  represents the surface boundary of the scene~\cite{Park_2019_CVPR}. While this formulation has advantages for scene decomposition, the multi-SDF design presents major drawbacks.

\subsection{Computational Overhead} 
To recover smooth surfaces with high-quality details, SDF-based learning methods often incorporate additional surface regularisation terms into the optimisation.
These typically include enforcing unit-norm gradients on the SDF function~\cite{gropp2020implicit} and constraining surface normals to vary smoothly over local neighbourhoods~\cite{Oechsle_2021_ICCV}.
Let $\mathcal{S}_{uni}$ be the set of points sampled uniformly across the 3D bounding space, and let $\mathcal{S}_{surf}$ be the set of points sampled in the vicinity of the scene surface $\mathcal{V}(\Omega)$ (refer to section~\ref{sec:details} for sampling details). Such regularisations are defined as:
\begin{align}
        \label{eq:regularisers_eik}
        \mathcal{L}_{eikonal} &= \frac{1}{N} \sum_{i=1}^N \frac{1}{\lvert \mathcal{S}^i_{uni} \rvert} \sum_{\textbf{x} \in \mathcal{S}^i_{uni}} (||\nabla\text{SDF}^i_{\theta}(\textbf{x})||_2 - 1)^2\,, \\
        \label{eq:regularisers_smo}
        \mathcal{L}_{smoothness} &= \frac{1}{N} \sum_{i=1}^N \frac{1}{\lvert \mathcal{S}^i_{surf} \rvert} \sum_{\textbf{x} \in \mathcal{S}^i_{surf}} ||\mathbf{n}^i(\textbf{x}) - \mathbf{n}^i(\textbf{x} + \varepsilon)||_2\,, 
    \end{align}
where $\mathbf{n}^i(\textbf{x})$ is the surface normal computed as $\mathbf{n}^i(\textbf{x}) = \nabla\text{SDF}^i_{\theta}(\textbf{x})/||\nabla\text{SDF}^i_{\theta}(\textbf{x})||_2$.
In a multi-SDF architecture, computing these regularisation terms requires evaluating every sampled point across all object SDFs, substantially increasing training time (see Table~\ref{fig:resources}).
This leads to a computational overhead that scales poorly with the number of objects in the scene.

\subsection{Gradient Discontinuities in Min-Pooling}
The $\min$ pooling operator provides a simple means of merging object-wise SDFs into a single scene representation. However, it is not a smooth function, introducing gradient discontinuities that cause unstable optimisation and slow convergence. 
Additionally, $\min$ pooling restricts information flow between adjacent objects, since each spatial location is assigned only to the SDF with the smallest value. 
This isolation prevents boundary regions from sharing gradients across objects, hindering joint refinement and invalidating the Eikonal constraint, which requires continuous gradients~\cite{Sitzmann2020ImplicitFunctions} (as shown by the fragmented chair legs in Figure~\ref{fig:replica_obj}).
In contrast, a single SDF representation leverages all available information, enabling more coherent learning across objects throughout the entire scene.

\begin{figure}[!ht]
    \centering

    \includegraphics[width=\linewidth]{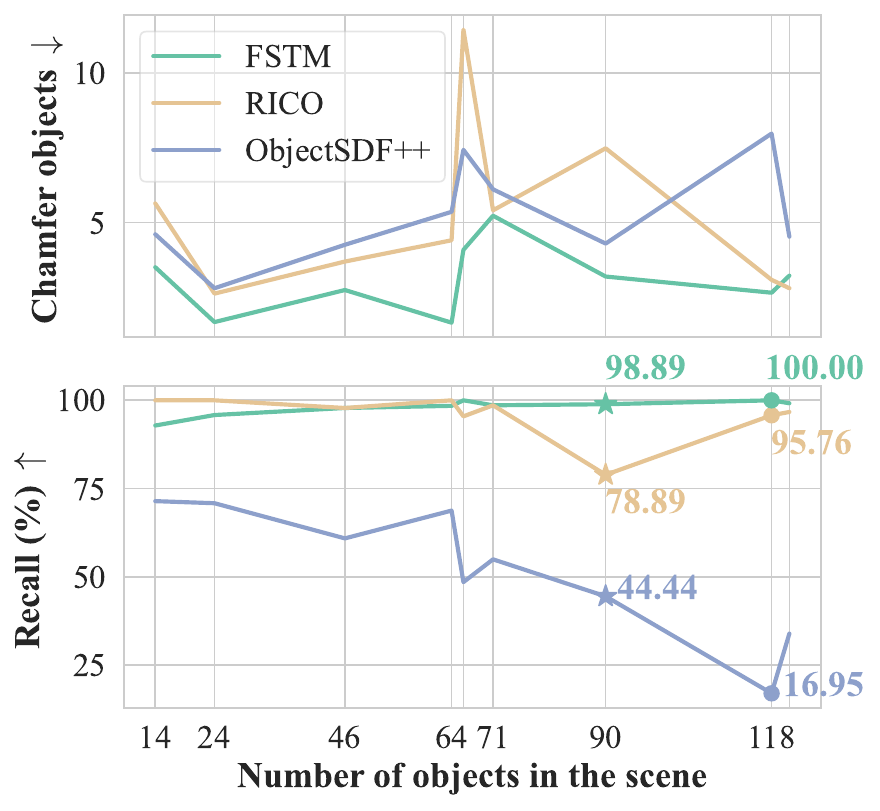}
    \caption{The optimisation of multi-SDF architectures degrades in complex scenes. In the plots we report scene complexity as the number of objects in the scene.
    Top: Chamfer Distance computed on the intersection of recovered objects. 
    Bottom: Comparison of object recall across methods. 
    The symbols $\star$ and $\bullet$ represent scene \textit{0a184cf634} and \textit{8be0cd3817} respectively.
    }
    \label{fig:objsdf_limitations}
\end{figure}

\subsection{Under-Observed Regions}
Multi-SDF architectures often struggle in under-observed regions of a scene. 
The object distinction loss introduced in~\cite{Wu_2023_ICCV} and the object point-SDF loss in~\cite{Li_2023_ICCV} aim to prevent unrealistic overlaps between objects by comparing their SDF values at nearby points. 
However, this constraint often results in coarse and oversimplified surface approximations in under-observed areas, leading to object boundaries that end abruptly instead of blending smoothly. 

In practice, these drawbacks result in quantifiable differences in reconstruction quality, as shown in Figure~\ref{fig:objsdf_limitations}. Our single SDF approach achieves a lower average distance between points on reconstructed surfaces and their corresponding ground-truth (i.e., Chamfer Distance) for the objects recovered by both methods (i.e., their intersection). Furthermore, the higher recall demonstrates that our method recovers a larger proportion of objects present in the scene. This benefit becomes increasingly significant with a growing number of objects in the scene, highlighting the scalability of our method.

\section{Method}
\label{sec:method}

\begin{figure*}[!t]
    \centering
    \begin{overpic}[width=\textwidth]{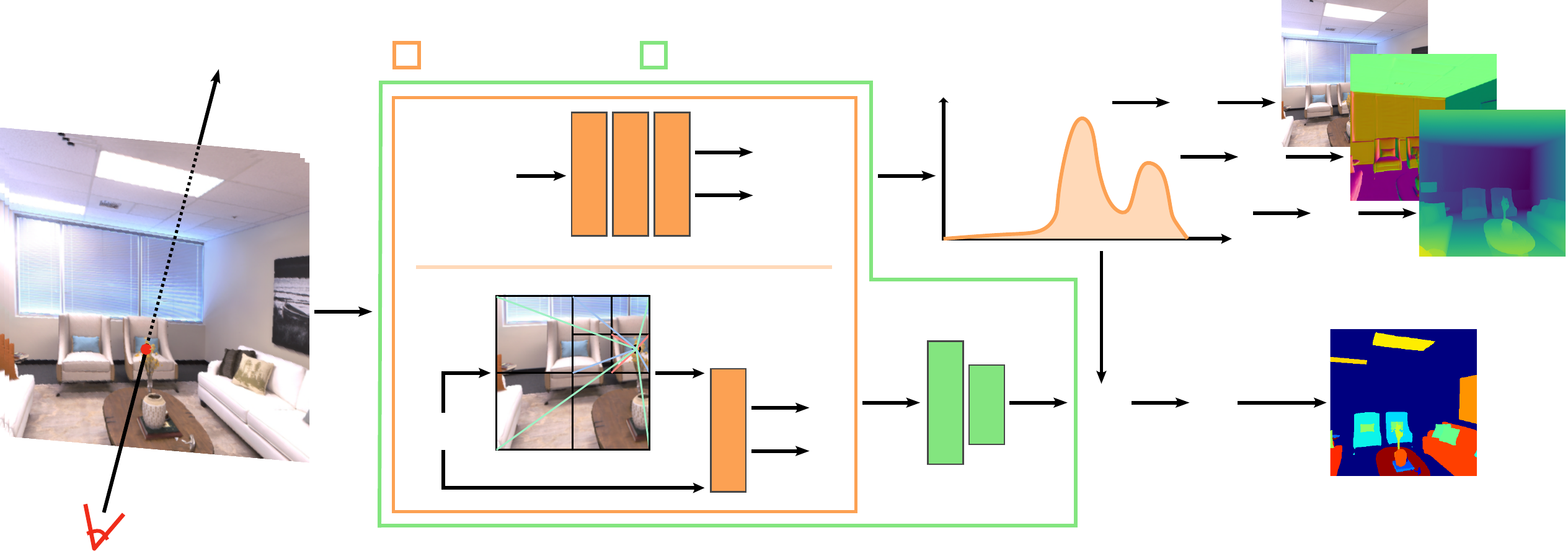}

        \put (125,310) { $r$}
        \put (775,180) { $t$}
        \put (730,100) {$\hat{S}^C$}

        \put (70,135) { $\mathbf{x}$}
        \put (270,72) { $\mathbf{x}$}
        \put (305,235) { $\mathbf{x}$}
        \put (270,310) { Step 1}
        \put (430,310) { Step 2}
        \put (250,260) { a)}
        \put (250,150) { b)}
        \put (480,48) { $\mathbf{g}$}
        \put (550,78) { $\mathbf{g}$}
        \put (445,210) { $\mathbf{g}$}
        \put (435,262) { $\text{SDF}_\theta$}
        \put (470,100) { $\text{SDF}_\theta$}
        \put (550,248) { $\text{SDF}_\theta$}
        \put (600,290) { $\sigma$}
        \put (645,100) { $\mathbf{s}^C$}
        \put (690,90) { $\int$}
        
        \put (758,86) { $\mathcal{L}_{semantic}$}
        \put (745,277) { $\mathcal{L}_{rgb}$}
        \put (788,242) { $\mathcal{L}_{normal}$}
        \put (833,205) { $\mathcal{L}_{depth}$}
        
    \end{overpic}
    \caption{FSTM framework. For clarity, the dependence on input variables is omitted. Panels a) and b) illustrate the two architectures: MLP and Multi-Res. Grids respectively.}
    \label{fig:framework}
\end{figure*}
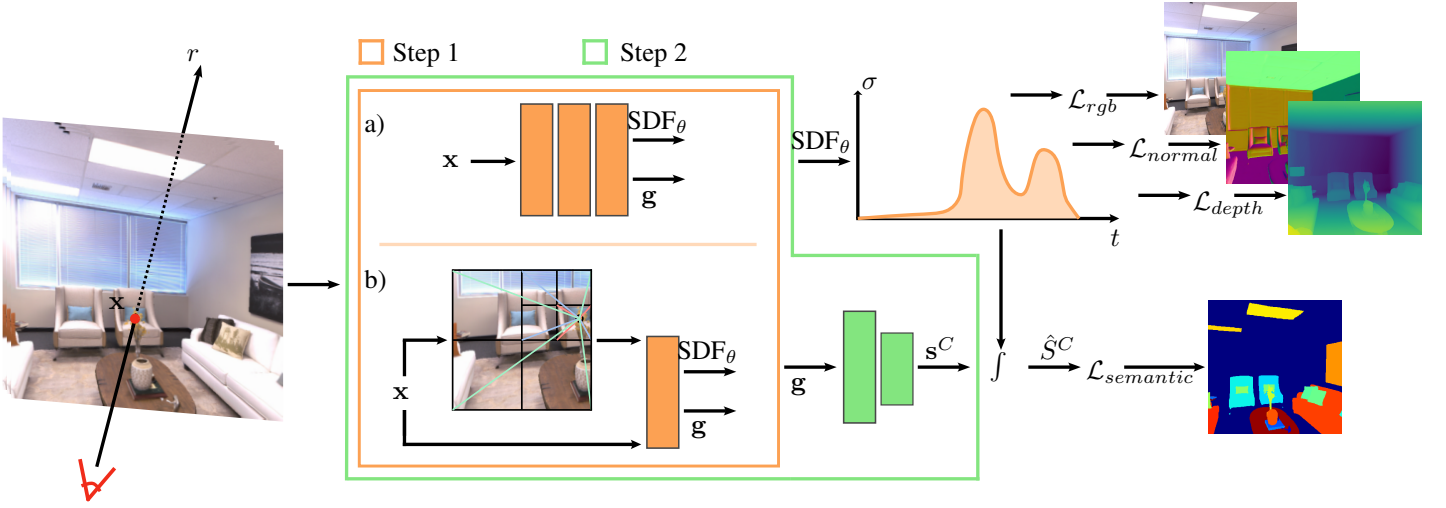


In this section, we introduce \NAME{} a streamlined framework for object reconstruction, illustrated in Figure~\ref{fig:framework}.
\NAME{} maintains generality by representing the entire scene with a single SDF,  while simultaneously enabling the estimation of a 3D semantic field. 

Distinct from recent work relying on multiple SDFs, our approach demonstrates that a single SDF model, optimised with a simple two-step training schedule, can achieve superior scene- and object-level reconstruction performance. Concretely, we decouple geometry and semantic learning into separate training phases, which leads to more reliable convergence and outperforms joint training and multi-SDF techniques in both accuracy and interpretability. 
At the end of training, a dedicated 3D segmentation post-processing step enables robust instance extraction from the reconstructed scene.

\subsection{Preliminaries}

Neural Radiance Fields (NeRF)~\cite{mildenhall_2021} represent a 3D scene as a continuous volumetric function mapping a spatial location $\mathbf{x} \in \mathbb{R}^3$ and a viewing direction $\mathbf{d} \in \mathbb{S}^2$ to a colour $\hat{c} \in [0,1]^3$ and volume density $\sigma \in \mathbb{R}_{\ge 0}$:
\begin{equation}
    F_\Phi: (\mathbf{x}, \mathbf{d}) \mapsto (\hat{c}, \sigma)\,,
\end{equation}
where $\Phi$ is parametrised by $\alpha$, $\beta$ and $\theta$.

For a camera ray $r(t) = \mathbf{o} + t \mathbf{d}$, the observed pixel colour is obtained via volume rendering:
\begin{align}
    \hat{C}(r) = \int_{t_n}^{t_f} T(t) \sigma(r(t)) \hat{c}(r(t), \mathbf{d}) dt\,, \label{eq:rendering}
    \\
    \text{where} \quad
    T(t) = \exp\Big(-\int_{t_n}^{t_f} \sigma(r(s)) ds \Big)
\end{align}
is the accumulated transmittance along the ray from near ($t_n$) to far ($t_f$).  
The photometric reconstruction loss is defined as:
\begin{equation}
    \mathcal{L}_{rgb} = \frac{1}{\lvert R \rvert} \sum_{r \in R} \|\hat{C}(r) - C(r)\|_1\,,
\end{equation}
where $R$ is the set of sampled rays and $C(r)$ is the ground-truth pixel colour.

Extensions of NeRF incorporate Signed Distance Functions (SDFs)~\cite{NEURIPS2022_9f0b1220,Wu_2023_ICCV} to model scene geometry explicitly.
The volume density $\sigma(\mathbf{x})$ is derived from the SDF function ($\text{SDF}_{\theta}$) via a differentiable mapping:
\begin{equation}
    \sigma(\mathbf{x}) = \alpha \Psi_\beta(-\text{SDF}_\theta(\mathbf{x}))\,,
\end{equation}
where $\Psi_\beta$ is the Cumulative Distribution Function (CDF) of the Laplace distribution with learnable scale $\beta$, and $\alpha$ and $\beta$ are optimised jointly with scene parameters~\cite{NEURIPS2021_25e2a30f}.


\subsection{\textbf{\NAME{}}}
Building on this foundation, \NAME{} represents the whole scene with a single learnable SDF, providing direct access to global geometry.

%
%
%
To enable joint geometry and semantic reconstruction, we extend MonoSDF~\cite{NEURIPS2022_9f0b1220} with a semantic decoder head. Because semantic labels are view-independent, the decoder is conditioned solely on spatial location. 
For a given 3D query point $\mathbf{x}$, the geometry MLP outputs both the SDF value and a feature vector $\mathbf{g}(\mathbf{x})$, which is mapped to $C$-class semantic logits $\mathbf{s}^C(\mathbf{x})$:
\begin{equation}
    \mathbf{x} \longmapsto \mathbf{g}(\mathbf{x}) \longmapsto \mathbf{s}^C(\mathbf{x},\mathbf{g}(\mathbf{x})),
\end{equation}
which are converted to probabilities via \textit{softmax}.
Following Eq.~\ref{eq:rendering}, the semantic distributions are integrated along each ray $r$:
\begin{equation}\label{eq:logits}
    \hat{S}^C(r) = \int T(t)\sigma(r(t))\mathbf{s}^C(r(t))dt,
\end{equation}
enabling dense per-ray semantic supervision.

\subsection{Optimisation Strategy}
To improve the SDF quality, we adopt both Eikonal and smoothness regularisation losses as defined in Eqs.~\ref{eq:regularisers_eik} and~\ref{eq:regularisers_smo} for $N=1$.
Following prior works, we leverage monocular geometric cues obtained from the pretrained Omnidata~\cite{Eftekhar_2021_ICCV} model, which predicts accurate normal and depth maps. This guidance enforces geometrically plausible reconstructions, especially in under-observed or ambiguous regions. 

Formally, the normal and depth supervision losses are defined as:
\begin{align}
    \mathcal{L}_{normal} &= 
    \frac{1}{\lvert R \rvert} \sum_{r \in R}||\hat{N}(r) - N(r)||_1  + ||1-\hat{N}(r)^T N(r)||_1 
    \\
    \mathcal{L}_{depth} &= \frac{1}{\lvert R \rvert} \sum_{r \in R}||(w\hat{D}(r) + q) - D(r)||_2 ,
\end{align}
where $N(r)$ and $D(r)$ denote the monocular normal and depth priors for ray $r$, and $\hat{N}(r)$ and $\hat{D}(r)$ are the corresponding volume-rendered predictions (cf. Eq.~\ref{eq:rendering}).
The scaling and shift factors $w$ and $q$ ensure scale- and shift-invariance of depth supervision, which is necessary due to the monocular depth ambiguity.

The semantic field $\mathbf{s}^C$ is optimised by minimising the cross-entropy loss between the predicted semantic distribution and the 2D ground-truth segmentation $p$:
\begin{equation}
    \mathcal{L}_{semantic} = - \mathbb{E}_{p}\left[\log\hat{S}^C(r)\right].
\end{equation}
%
The overall training objective minimises the weighted sum:
\begin{multline}\label{eqn:finalBoss}
    \mathcal{L} = \mathcal{L}_{rgb} + \lambda_1\mathcal{L}_{eikonal} + \lambda_2\mathcal{L}_{smoothness} ~+ \\ \lambda_3\mathcal{L}_{normal} + \lambda_4\mathcal{L}_{depth} + \lambda_5\mathcal{L}_{semantic}.
\end{multline}

\subsection{2-Step Training}
To stabilise the optimisation process and improve the final reconstruction quality, we adopt a two-stage training schedule. To ensure fair comparisons and maintain generality, we adhere to the standard training budget established by MonoSDF~\cite{NEURIPS2022_9f0b1220}, allocating for each stage half of the iterations.
During this initial phase, the geometry network is trained independently by setting $\lambda_5=0$ in Eq.~\ref{eqn:finalBoss}. 
This warm-up phase ensures convergence of scene geometry and prevents early semantic supervision from degrading reconstruction quality, as observed empirically in Table~\ref{tab:replica_shrinked}.
Early utilisation of 2D semantic supervision can bias the geometry network to prioritise semantic consistency over geometry accuracy, resulting in suboptimal local minima that are difficult to recover from. 

In the second stage, we jointly optimise both geometry and semantics for the remainder of the training budget. While the exact iteration split could be dynamically tuned based on scene complexity or dataset characteristics, we maintain a fixed equal division to prioritise reproducibility and avoid scene-specific hyperparameter tuning, demonstrating robust performance across diverse synthetic and real-world datasets.

\subsection{3D Segmentation}
To obtain object-level segmentations at test time, we draw inspiration from 3D texturing techniques~\cite{Yu2022SDFStudio}. 
Specifically, the reconstructed surface mesh $\mathcal{M} = \{F_i\}_{i=1}^M$ is obtained by applying an iso-surface extraction algorithm (e.g., Marching Cubes) to the zero-level set of the learned single SDF. For each mesh face $F_i$, we compute its centre $\mathbf{x}_i \in \mathbb{R}^3$ and corresponding surface normal $\mathbf{n}_i$. 

We define a ray $r_i = (\mathbf{o}_i, \mathbf{d}_i)$ as follows:
\begin{equation}
\mathbf{o}_i = \mathbf{x}_i + \epsilon \mathbf{n}_i, \quad \mathbf{d}_i = -\mathbf{n}_i,    
\end{equation}
where $\epsilon$ is a small offset distance along the normal direction, positioning the ray origin just in front of the face centre. 
To evaluate the semantic distribution $\hat{S}^C(r_i)$ for the face, we integrate the semantic logits along $r_i$ using Eq.~\ref{eq:logits}. The integration is performed over the ray distance interval $t \in [0, \epsilon]$, effectively accumulating predictions from the offset origin through the surface intersection.
The semantic label for face $F_i$ is assigned by selecting the class with the highest predicted probability:
\begin{equation}
c_i = \underset{k \in \{1, \ldots, C\}}{\mathrm{argmax}} \hat{S}^C(r_i)_k.
\end{equation}
This approach efficiently produces dense, object-wise semantic labels on the reconstructed surface, leveraging GPU acceleration for practical scalability.


\subsection{Implementation Details}\label{sec:details}
Our implementation is based on PyTorch~\cite{pytorch} and leverages the Adam optimiser~\cite{kingmaAdam}.
For the geometry network, we evaluate two architectures from MonoSDF ~\cite{NEURIPS2022_9f0b1220}: 
(i) a Multi-Resolution Feature Grid (Multi-Res-Grids) combined with a shallow MLP, and (ii) a deeper standalone MLP (MLP). 
Consistent with prior findings, the Multi-Res-Grids backbone achieves better performance on synthetic datasets, while the deep MLP prevails on real-world datasets. Our experiments present results for these configurations in Sections~\ref{subsec:replica} and~\ref{subsec:scannetpp} accordingly.
The semantic network follows the Semantic-NeRF~\cite{Zhi_2021_ICCV} design, adapted to accept the 256-dimensional geometry feature vector $\mathbf{g}(\mathbf{x})$.

Learning rates are set to $5\times 10^{-4}$ for both semantic and geometry networks, and $1\times 10^{-2}$ for the Multi-Res-Grids.
We follow the training schedule proposed in MonoSDF, using 200k iterations for Replica and 400k iterations for ScanNet++ for fair comparison.

During the geometry warm-up, hyperparameters $\lambda_{1-4}$ are set to $0.1$, $0.005$, $0.05$, and $0.1$, respectively.
For the joint training step, we reduce $\lambda_{2-4}$ by a factor of 10, while keeping $\lambda_1$ unchanged, and set $\lambda_5 = 0.1$ for semantic supervision.
On real-world datasets, we swap the values of $\lambda_3$ and $\lambda_4$, as we observe greater numerical stability from monocular normal estimation. 

To evaluate the surface smoothness regularisation presented in Eq.~\ref{eq:regularisers_smo}, points must be sampled in the immediate vicinity of the scene geometry. Rather than relying on purely uniform sampling, we leverage the iterative \textit{ErrorBoundSampler} introduced in VolSDF~\cite{NEURIPS2021_25e2a30f}. This sampling strategy naturally concentrates point evaluations at the peaks of the volumetric density distribution, guaranteeing that the smoothness constraint is strictly enforced near the SDF zero-level set.

Finally, to determine the ray origin offset $\epsilon$ for the 3D segmentation probing, we dynamically scale the offset to the resolution of the reconstructed geometry by setting $\epsilon$ equal to the mean edge length of the extracted triangular mesh.

\section{Experiments}
\label{sec:experiments}




\subsection{Datasets}
We evaluate \NAME{} on two challenging indoor surface reconstruction benchmarks, both featuring complex scenes with multiple objects. 
The first is the Replica dataset~\cite{replica19arxiv}, a synthetic benchmark providing high-quality ground-truth geometry and semantic annotations. 
The second is ScanNet++~\cite{yeshwanthliu2023scannetpp}, a large-scale real-world dataset with complex indoor environments, making it well-suited for assessing the semantic supervision role in the absence of reliable monocular estimates.


\subsection{Baselines}
We compare our method against five state-of-the-art methods.
MonoSDF~\cite{NEURIPS2022_9f0b1220} serves as a strong single-SDF baseline for evaluating scene-level accuracy.
ObjectSDF++~\cite{Wu_2023_ICCV}, RICO~\cite{Li_2023_ICCV}, and PhyRecon~\cite{NEURIPS2024_2d880acd} represent multi-SDF methods that enhance object reconstruction accuracy.
For PhyRecon, as the original training pipeline is not readily reproducible, we limit our comparison to the pre-computed meshes provided by the authors for the Replica dataset. Since we follow the semantic class mapping of ObjectSDF++, and PhyRecon does not provide all object-level meshes, in our experiments, we only compare scene-level accuracy and one qualitative example.
Finally, we compare against ObjectGS~\cite{zhu2025objectgs} in the controlled Replica dataset. Since ObjectGS requires semantic labels on an initial 3D point cloud, we sample 50k points from ground-truth meshes and assign semantics via exact ray-mesh intersection using known camera poses. 
While this initialisation is more favourable than its original setting, which relies on noisy, unprovided SfM points, we adopt this setup to transparently assess the upper-bound potential of ObjectGS under these specific dataset constraints.



\subsection{Metrics}
Following previous works~\cite{NEURIPS2022_9f0b1220,Wu_2023_ICCV,Guo_2022_CVPR}, we evaluate reconstruction quality using Chamfer distance, Normal Consistency, and F-score at a $5 cm$ threshold. 
For object-level evaluations, we also compute the 95th percentile Hausdorff distance, which captures the largest discrepancies in minimum point-to-surface distances while ignoring extreme outliers.


\subsection{Reconstructions on Replica}\label{subsec:replica}

The Replica dataset features high-quality synthetic data, making it ideal for evaluating the reconstruction accuracy of NeRF-based methods under controlled settings.
Following prior work~\cite{Zhi_2021_ICCV,NEURIPS2022_9f0b1220,Wu_2023_ICCV}, we select eight commonly used scenes for our experiments. 

\input{tables/replica_shrinked}

\begin{figure*}[!ht]
\centering
\begin{overpic}[width=\textwidth]{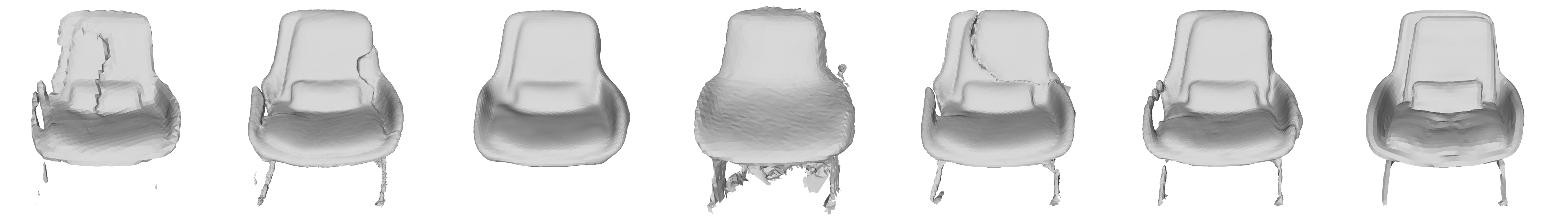} 
    

    \put (0,50) {\rotatebox{90}{\normalsize \textit{office4}}}
    
    \put (15,-20) {\normalsize ObjectSDF++}
    \put (165,-20) {\normalsize PhyRecon}
    \put (330,-20) {\normalsize RICO}
    \put (460,-20) {\normalsize ObjectGS}
    \put (590,-35) 
        {\normalsize \shortstack{Joint\\optimisation}}
    \put (760,-20) {\normalsize FSTM}
    \put (915,-20) {\normalsize GT}
    
\end{overpic}
\vspace{5pt}
\caption{Qualitative evaluation of an object-level reconstruction on the Replica dataset. 
Our two-step method more accurately approximates the true surface, capturing finer geometric details, such as the legs of the chair in \textit{office4}.}
\label{fig:replica_obj}
\end{figure*}
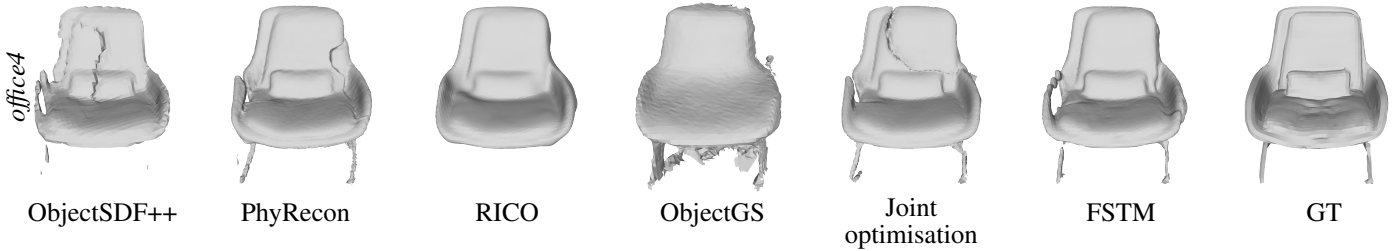



We evaluate reconstruction performance at both the scene and object levels to assess the benefit of semantic supervision. 
As shown in Table~\ref{tab:replica_shrinked}, naive joint optimisation yields lower performance compared to the MonoSDF baseline, which does not incorporate semantic information. This suggests that directly optimising geometry and semantics jointly can lead to suboptimal convergence. 
In contrast, our proposed two-step training strategy consistently achieves the best results across most scenes and metrics. 
A similar trend is observed at the object level, where our approach yields the most accurate reconstructions overall, followed by RICO. Interestingly, while joint optimisation performs worse at the scene level, it slightly outperforms ObjectSDF++ at the object level, although it remains less stable across scenes. In comparison, ObjectGS also underperforms FSTM at both the scene and object levels, exhibiting higher Chamfer errors and reduced geometric fidelity.

Qualitative analysis further illustrates the impact of semantic supervision on object-level reconstruction. 
As illustrated in Figure~\ref{fig:replica_obj}, surface artefacts often arise from joint optimisation and persist throughout training. 
For instance, in the reconstruction of the chair object in \textit{office4}, both the naive approach and ObjectSDF++ exhibit pronounced errors, whereas our method achieves notably finer details and smoother surfaces. Although RICO produces visually smooth reconstructions, it often fails to capture finer structural elements such as chair legs, aligning with findings reported in previous work~\cite{NEURIPS2024_2d880acd}.
Despite the favourable experimental setting, ObjectGS performs poorly on this object, likely due to its explicit Gaussian representation, which is often limited in recovering topologically consistent 3D surfaces~\cite{li2024monogsdf}.
These qualitative improvements are further supported by higher Normal Consistency scores, indicating smoother reconstructed surfaces. Quantitatively, \NAME{} produces more accurate geometry, with Chamfer and Hausdorff distances improving by up to $24\%$ across the dataset.


\input{tables/scannetpp_shrinked}

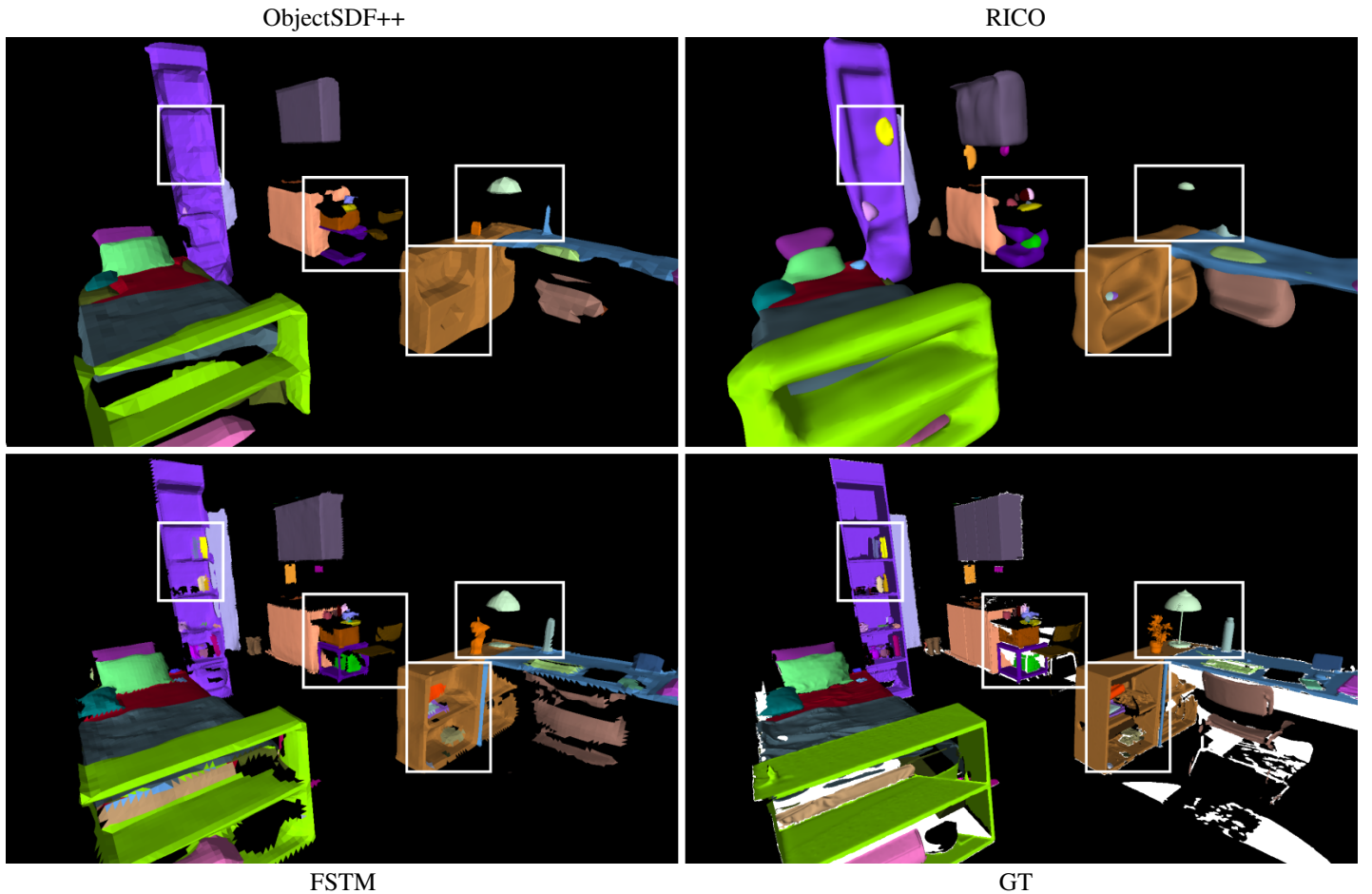
\begin{figure*}[!ht]
\vspace*{1em}
\centering
\begin{overpic}[width=\textwidth,percent]{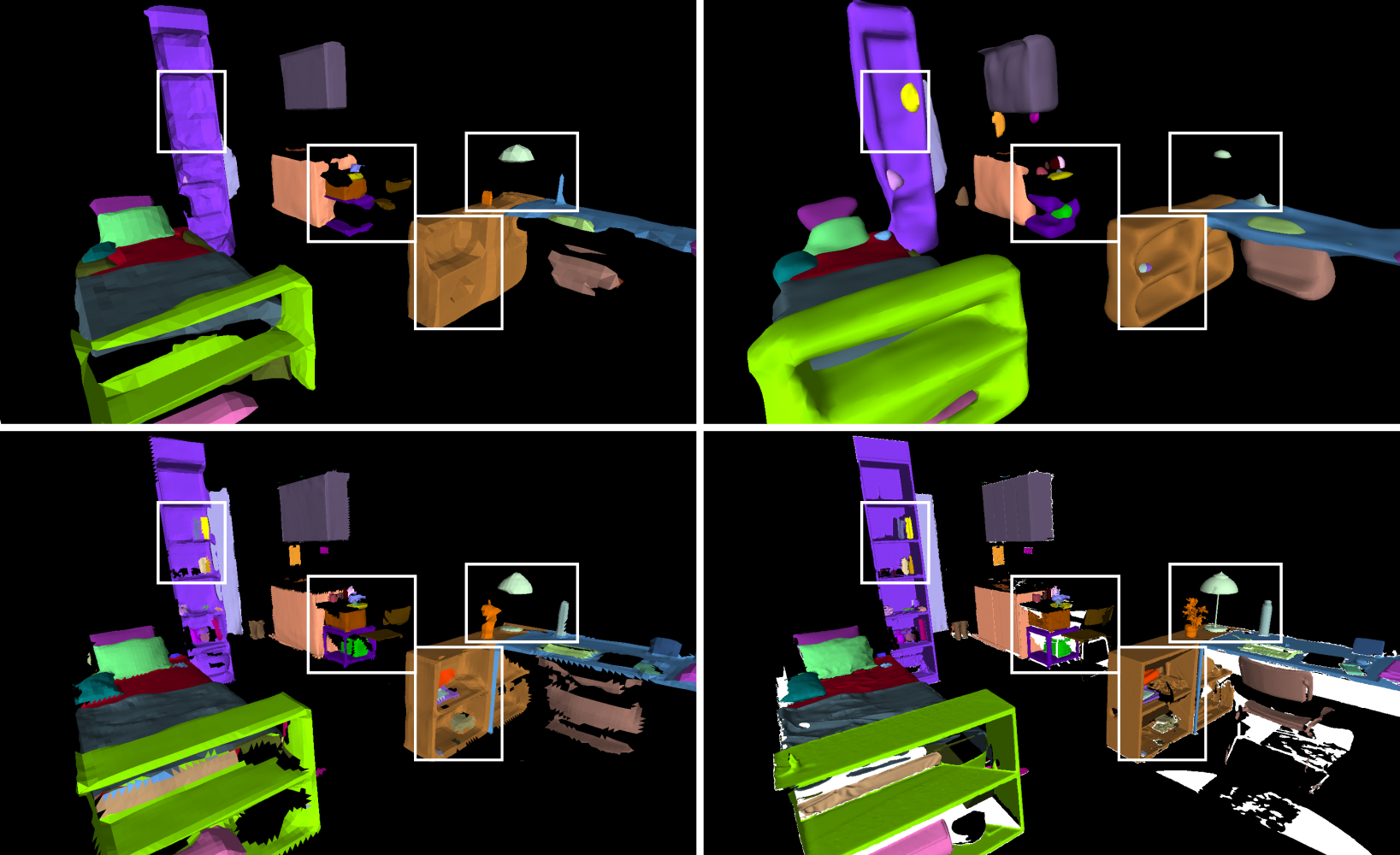}
    \put(19,62){\normalsize ObjectSDF++}
    \put(72.5,62){\normalsize RICO}
    \put(22.5,-2){\normalsize FSTM}
    \put(73.5,-2){\normalsize GT}
\end{overpic}
\vspace{0pt}
\caption{Qualitative comparison of 3D object segmentation for scene \textit{0a184cf634} from the ScanNet++ dataset, which contains 90 objects. Both ObjectSDF++ and RICO struggle to identify small objects (e.g., items on the shelves) and tend to oversimplify unobserved regions of the reconstruction (e.g., the cabinet structure).}
\label{fig:scannet_qualitative_0a}
\end{figure*}

\subsection{Reconstructions on ScanNet++}
\label{subsec:scannetpp}


To evaluate the real-world applicability of our method, we conduct experiments on the ScanNet++ dataset. 
Unlike Replica, ScanNet++ presents practical challenges, such as imprecise camera poses and blurry frames.
%
Since ScanNet++ was released after the compared baselines were published, we established an evaluation subset by initially sampling 12 scenes covering a wide range of object counts (ranging from a few dozen to over one hundred). To ensure a fair and direct comparison across all metrics, we applied a strict inclusion rule: a scene was only kept if all evaluated methods successfully converged. Consequently, 2 scenes were excluded due to ObjectSDF++ failing to access object surfaces after 20k iterations (e.g., scene \textit{e3e0617f98}, a complex environment with over 200 objects and 400 training views). This filtering resulted in the final 10 representative scans used in our experiments, which were preprocessed using the same pipeline as Replica.

As with Replica, we evaluate both scene-level and object-level reconstruction performance (Table~\ref{tab:scannetpp_shrinked}) and present qualitative 3D segmentation results (Figure~\ref{fig:scannet_qualitative_0a}). 
Our method achieves state-of-the-art performance on real-world data, maintaining strong results across scenes of varying complexity. 
To ensure fairness, we report object-level metrics computed only over the intersection of recovered objects across the methods.

Additionally, consistent with the analysis in Figure~\ref{fig:objsdf_limitations}, our method achieves higher object recall, particularly in scenes with dense object distributions, as scene \textit{0a184cf634} illustrated in Figure~\ref{fig:scannet_qualitative_0a}.
An exception is noted in scene \textit{0271889ec0}, where, despite recovering six times as many objects as ObjectSDF++ and four times as many as RICO, the presence of a single outlier skewed the average object-level metrics.

\input{tables/resources_small_single_column}

\begin{figure}
    \centering
    \includegraphics[width=\linewidth]{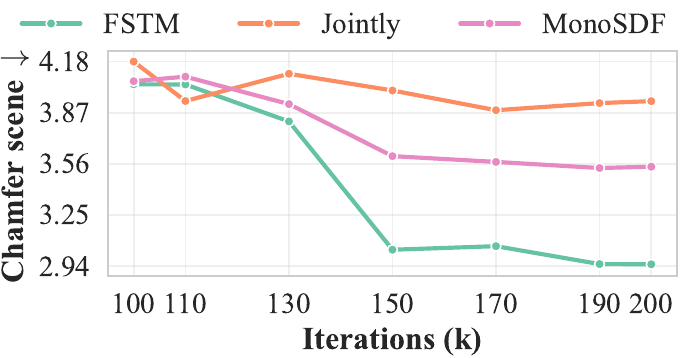}
    \caption{Convergence behaviour across the 8 Replica scenes from mid-training. FSTM shows a steep curve after the optimisation switches from the initial semantic warm-up stage of 10k iterations to joint fine-tuning.}
    \label{fig:convergence}
\end{figure}



\subsection{Computational Complexity}

We compare the time and memory complexity of our method on the Replica dataset using the Multi-Res-Grids architecture. 
Each scan is composed of 100 images with comparable scene complexity, allowing us to isolate computational differences across methods. 
For a fair comparison, we follow the default MonoSDF training schedule and fix the training budget to 200k iterations for all methods evaluated on Replica. 
This controlled setting avoids confounding factors commonly present in real-world datasets, where scene scale and image count may vary significantly under a fixed iteration budget.

A key advantage of our method is its compatibility with any single-SDF framework, requiring only the addition of a semantic decoder head. 
As shown in Table~\ref{fig:resources}, this design results in minimal computational overhead compared to our backbone (MonoSDF). In contrast, both ObjectSDF++ and RICO exhibit runtimes more than 2.3$\times$ slower.

We further analyse how properly integrated semantic priors accelerate and improve geometric optimisation. Figure~\ref{fig:convergence} tracks scene accuracy starting from 100k iterations, the exact point where our method introduces the semantic signal. 
The plot suggests that while naive joint optimisation irreversibly damages geometric fidelity, our two-stage \NAME{} approach uses semantics to actively improve the geometry, ultimately surpassing the final accuracy of the semantic-free MonoSDF baseline. Furthermore, this rapid convergence demonstrates that the semantic head requires significantly fewer iterations to stabilise. Consequently, in time-constrained scenarios, the second training schedule can be safely reduced by 50k iterations. This allows users to effectively reduce the total training time by $\sim$25\% while still achieving superior geometric reconstruction compared to the full-length baseline.


\begin{figure*}[!ht]

    \makebox[\linewidth][c]{%
        \hspace*{0.01\linewidth}
        \begin{overpic}[width=0.96\linewidth,percent]{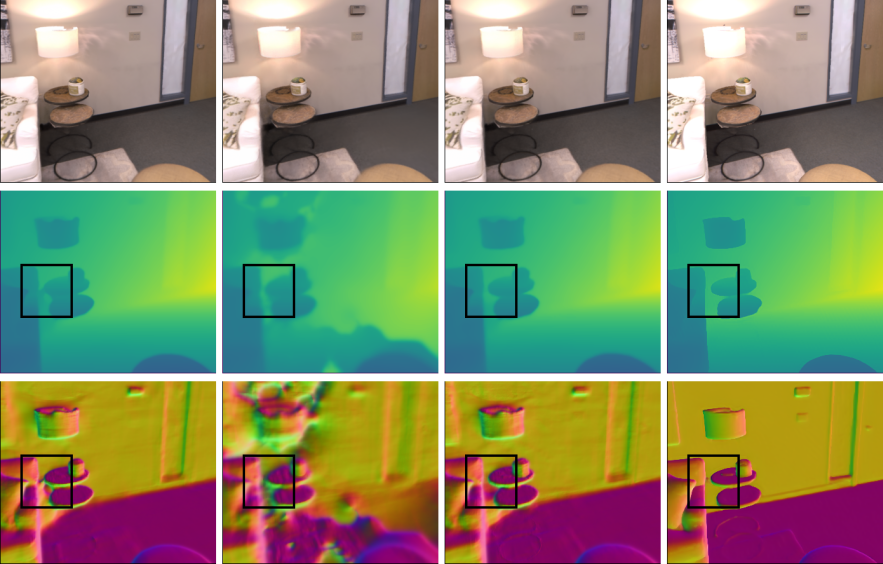}
            \put(12,-2){\normalsize (a)}
            \put(37,-2){\normalsize (b)}
            \put(62,-2){\normalsize (c)}
            \put(87,-2){\normalsize GT}

            \put(-2,4){\rotatebox{90}{\normalsize Rendered RGB}}
            \put(-2,25){\rotatebox{90}{\normalsize Rendered Depth}}
            \put(-2,46){\rotatebox{90}{\normalsize Rendered Normals}}
        \end{overpic}%
    }




    \vspace{10pt}
    \caption{Colour, depth and normal map novel-view renderings for Replica scene \textit{room0}. (a) Reconstruction without geometric or semantic priors ($\lambda_{3,4,5} = 0$); (b) and (c) reconstructions with semantic priors enabled but geometric priors disabled ($\lambda_{3,4} = 0$, $\lambda_5 = 0.1$), where (b) shows joint optimisation and (c) the two-step approach. 
    The highlighted area demonstrates that semantic priors, when applied within our two-step training approach, effectively enhance the capture of geometric details, unlike those same priors under naive joint optimisation.
    }
    \label{fig:sem_ablation}
\end{figure*}
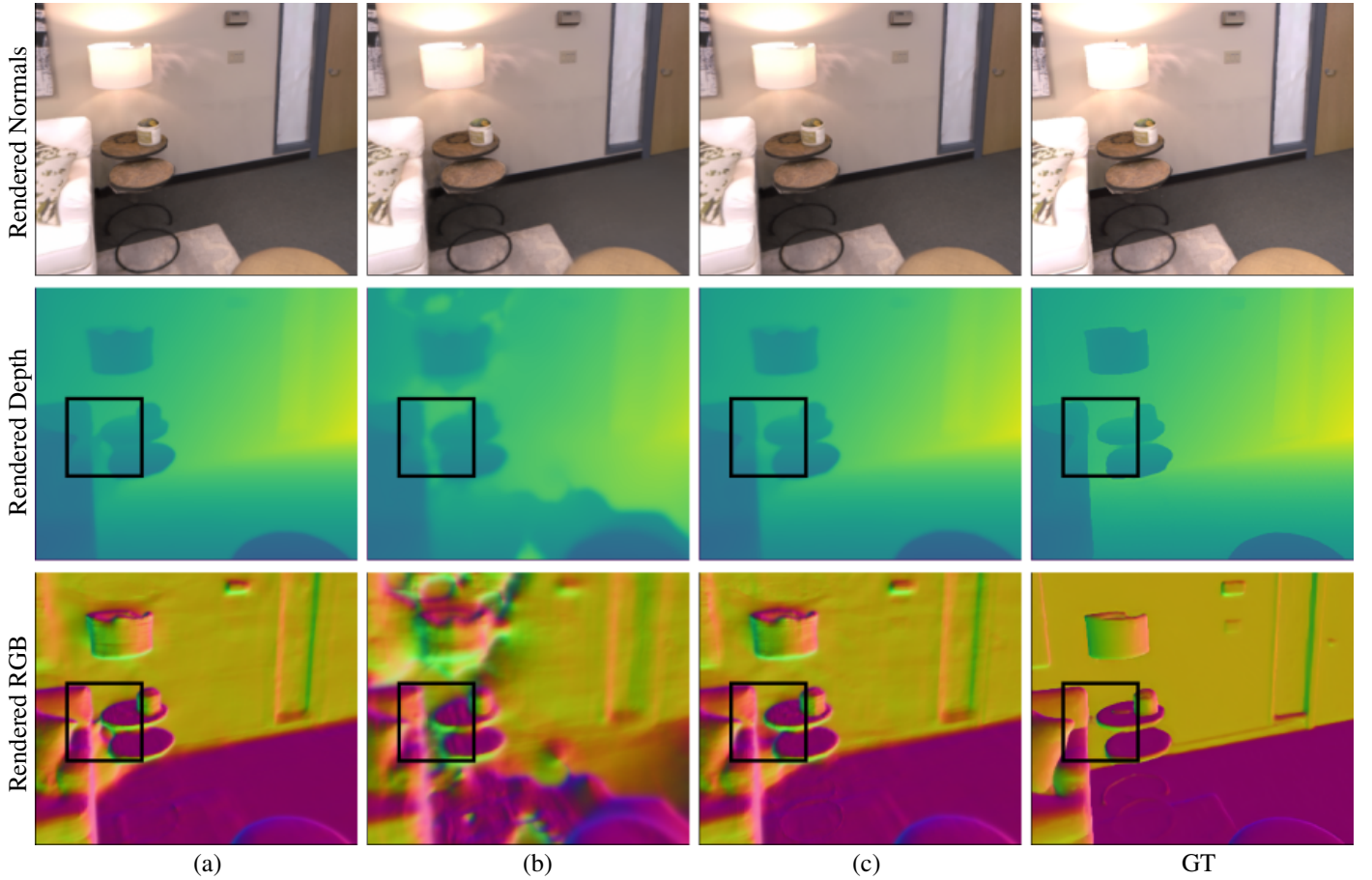

\subsection{Effect of Semantic Supervision on Scene Geometry}
To illustrate the role of semantic supervision, we perform a qualitative experiment on the Replica dataset. We compare three configurations of the \NAME{} model: 
one trained without semantic or geometric priors (i.e., $\lambda_{3,4,5} = 0$), and two others trained with only semantic priors enabled (i.e., $\lambda_{3,4} = 0$, $\lambda_5 = 0.1$) in both joint optimisation and two-step settings. 
We visualise the rendered geometric properties of the reconstructed scenes alongside the colour renderings. 
As shown in Figure~\ref{fig:sem_ablation}, semantic priors encourage the learning of 3D structure even without explicit geometric supervision.
Furthermore, object boundaries become more clearly defined, resulting in a more faithful reconstruction of scene geometry. However, these improvements are only achieved with our two-step approach and are absent under naive joint optimisation.

\begin{figure}[!ht]
    \centering
    \begin{overpic}[width=\linewidth]{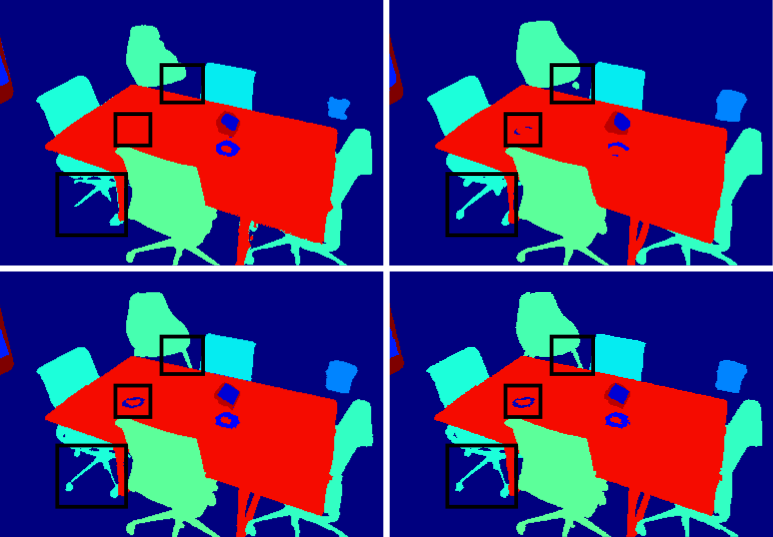}
    \put(155,705) {\normalsize ObjectSDF++}
    \put(710,705) {\normalsize RICO}
    \put(200,-35) {\normalsize FSTM}
    \put(740,-35) {\normalsize GT}
    \end{overpic}
    \vspace{-5pt}
    \caption{Novel-view 2D segmentation renderings for Replica scene \textit{office2}. Highlighted regions depict segmentation detail lost in multi-SDF architectures.}
    \label{fig:seg_qualitative}
\end{figure}
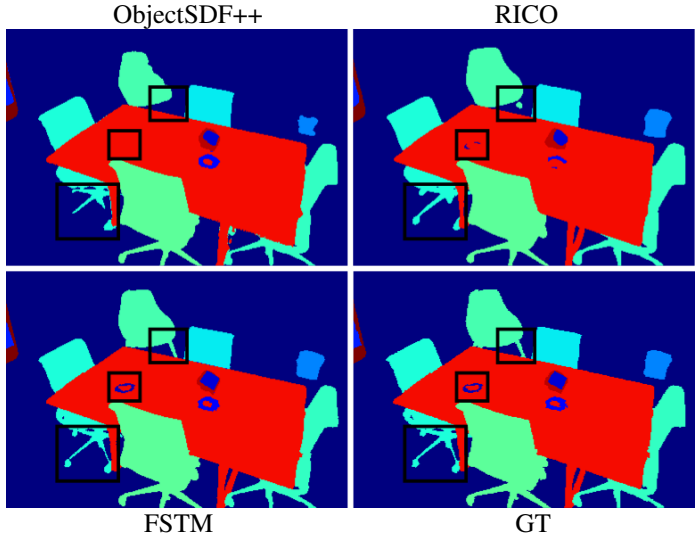

\subsection{2D Segmentation Performances}
We evaluate the compared methods in terms of their 2D segmentation performance on novel views using the Replica dataset.
Qualitative results are presented in Figure~\ref{fig:seg_qualitative} and quantitative metrics are reported in Table~\ref{tab:2d_seg}. Across both evaluations, \NAME{} outperforms the compared methods, supporting earlier findings. Notably, our approach successfully reconstructs fine geometric details that are often lost or over-smoothed in multi-SDF architectures. Overall, while colour renderings remain quite consistent across the methods, the mean Intersection over Union (mIoU) reveals a substantial improvement in 2D novel-views segmentation with our method.

\input{tables/segmentation_2D}

\subsection{Limited Supervision}

\begin{figure}[!ht]
    \centering
    \includegraphics[width=\linewidth]{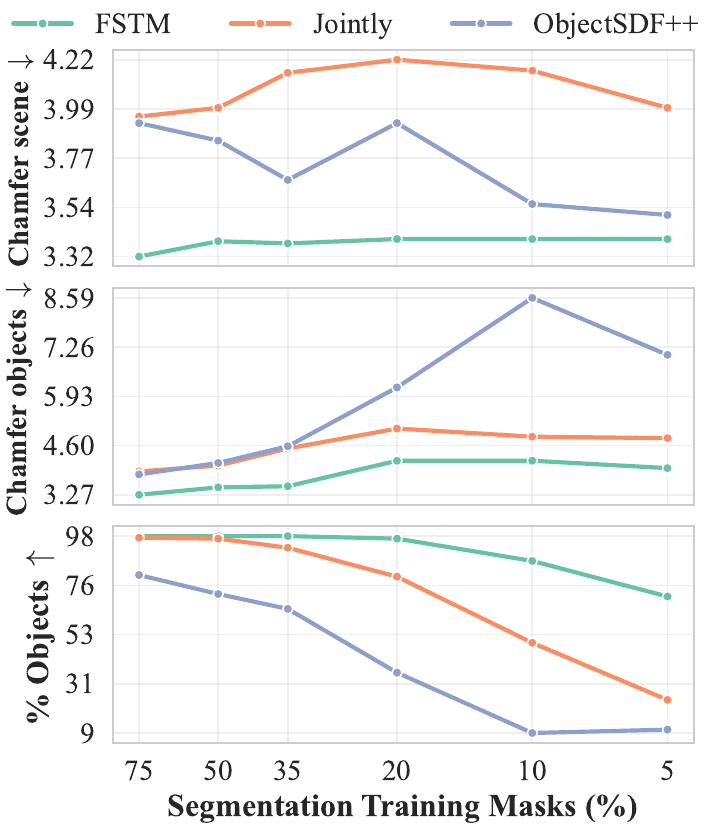}
    \caption{
    Evaluation under sparse semantic supervision across the 8 Replica scenes. Results are shown for scene-level reconstruction quality, object-level performance, and object recovery. FSTM maintains stable scene reconstruction quality and demonstrates robustness in object recovery.}
    \label{fig:sparse}
\end{figure}

To evaluate more realistic settings with limited dense supervision, Figure~\ref{fig:sparse} shows results obtained using several subsets of the available segmentation masks for training. The scene-level plot reveals that joint optimisation and ObjectSDF++ improve (i.e., Chamfer Distance decreases) as supervision becomes sparser. This behaviour suggests that dense semantic supervision can adversely affect geometric optimisation when trained naively or under a multi-SDF architecture. By contrast, the proposed two-step formulation maintains near-constant scene-level reconstruction quality, degrading only gradually as supervision becomes sparse, as expected.

The object-level results require more careful interpretation. While \NAME{} achieves the best performance overall, the apparent gain at 5\% supervision must be considered jointly with the object recovery rate shown in the plot at the bottom. For fairness, object-level metrics are computed only over the intersection of objects reconstructed by all compared methods. At 5\% supervision, \NAME{} still recovers nearly 75\% of all objects, whereas ObjectSDF++ recovers only 10\%. Consequently, the common evaluation set becomes extremely small for most Replica scenes. 
Nevertheless, \NAME{} achieves the best performance across all supervision levels.

\subsection{Limitations}
While \NAME{} effectively leverages semantic priors to improve implicit surface reconstruction, we identify a few limitations inherent to our design choices. 

First, because our method reconstructs a single global SDF manifold and derives object instances via semantic labelling, it does not natively output independent, watertight meshes for each object. Consequently, in heavily occluded or unobserved regions, the extracted object surfaces may contain holes. While multi-SDF formulations extrapolate unseen regions to force closed surfaces, visibility remains a fundamental limitation of non-generative reconstruction. For downstream applications strictly requiring closed geometries (e.g., 3D printing), standard post-processing algorithms (such as MeshFix~\cite{attene2010lightweight}) must be applied to our outputs.

Second, the geometric enhancements provided by our semantic head depend on a minimum density of semantic observations. Our ablation studies demonstrate that \NAME{} remains robust under sparse semantic supervision (e.g., dropping from 100 to 20 views). However, as the availability of 2D semantic annotations becomes severely limited (e.g., 5 to 10 views), the scene-level geometric improvements over the semantic-free MonoSDF baseline become increasingly marginal.

Finally, \NAME{} is a per-scene optimisation pipeline. Unlike recent feed-forward 3D reconstruction networks~\cite{tang2025mv,ren2025fin3r} that prioritise rapid inference or open-vocabulary editability, our method cannot achieve real-time reconstruction. Although our two-stage approach significantly accelerates convergence compared to naive joint training, the overall wall-clock time remains higher than 3D Gaussian Splatting (3DGS) approaches. We accept this trade-off, as per-scene SDF optimisation remains critical for domains requiring high-fidelity, topologically correct surfaces, such as industrial inspection and surgical planning, where current rapid-inference methods still struggle with geometric accuracy.

\section{Conclusion}
\label{sec:conclusion}
%
This work presents \NAME{}, a unified framework that integrates geometry and semantics within a single implicit surface representation. Rather than treating semantics as a mere byproduct of scene reconstruction, we demonstrate how properly scheduled semantic supervision actively refines and recovers high-fidelity geometry. By introducing a two-stage optimisation schedule, \NAME{} addresses the destructive interference caused by early semantic gradients, ensuring robust 3D reconstruction across both synthetic and real-world datasets. Furthermore, by maintaining a single global SDF, our approach avoids gradient discontinuities and severe computational overhead inherent to multi-SDF designs. Ultimately, \NAME{} delivers improved geometric accuracy with superior efficiency, establishing a streamlined foundation for scalable, semantically informed 3D scene understanding.

\bibliographystyle{IEEEtran}
\bibliography{main}  

\end{document}

%% file: tables/replica_shrinked.tex
\begin{table*}[!t]
    \centering
    \caption{Scene-level and object-level reconstruction results average across the Replica dataset using Multi-Res-Grids architecture. 
    Our 2-step method shows consistent improvements among all the scans.}
    \large
     \begin{tabular}{@{}l ccc ccc@{}}
        \toprule

        \multirow{2}{*}{\textbf{Methods}} &\multicolumn{3}{c}{\textbf{Scene-level}} &\multicolumn{3}{c}{\textbf{Object-level}} \\
        \cmidrule(lr){2-4}
        \cmidrule(lr){5-7}

        \multicolumn{1}{c}{} &\textbf{Normal C.}$\uparrow$ &\textbf{Chamfer-L1}$\downarrow$ &\textbf{F-Score}$\uparrow$ &\textbf{HD-95}$\downarrow$ &\textbf{Chamfer-L1}$\downarrow$ &\textbf{F-Score}$\uparrow$ \\
        \midrule
    
        MonoSDF & 89.41 & 3.59 & 85.84 &--&--&-- \\
        ObjectSDF++ & 90.30 & 3.58 & 86.14 &5.34&3.78&79.79 \\
        PhyRecon & 86.59 & 5.46 & 78.33 &--&--&-- \\
        RICO & 89.54 & 4.36 & 82.98 & 4.54 & 3.35 & 83.02 \\
        ObjectGS &90.79&4.43&77.99&4.49&3.49&79.78 \\
        Joint optimisation & 90.34 & 3.93 & 85.41 &5.07&3.58&81.40 \\
        FSTM & \textbf{92.02} & \textbf{2.94} & \textbf{89.38} &\textbf{4.02}&\textbf{2.86}&\textbf{85.86} \\
        
        \bottomrule
    \end{tabular}
    
    
    \label{tab:replica_shrinked}
\end{table*}

%% file: tables/scannetpp_shrinked.tex
\begin{table*}[!t]
    \caption{Evaluation of both scene and object-level reconstruction for the ScanNet++ dataset using the MLP configurations. We considered a budget of 400k iterations for both methods, double the default, to allow complex scenes to plateau.}
    \centering
    \large
    \begin{tabular}{@{}l ccc ccc@{}}
        \toprule

        \multirow{2}{*}{\textbf{Methods}} &\multicolumn{3}{c}{\textbf{Scene-Level}} &\multicolumn{3}{c}{\textbf{Object-Level}} \\
        \cmidrule(lr){2-4}
        \cmidrule(lr){5-7}
        
        \multicolumn{1}{c}{} &\textbf{Normal C.}$\uparrow$ &\textbf{Chamfer-L1}$\downarrow$ &\textbf{F-Score}$\uparrow$ &\textbf{HD-95}$\downarrow$ &\textbf{Chamfer-L1}$\downarrow$ &\textbf{F-Score}$\uparrow$ \\
        \midrule
    
        ObjectSDF++ & 84.68 & 6.59 & 70.19 & 8.32 & 5.56 & 72.93 \\
        RICO &80.22&10.98&60.57 & 9.23&6.17&69.32 \\
        FSTM & \textbf{86.74} & \textbf{5.00} & \textbf{76.54} & \textbf{5.38} & \textbf{3.78} & \textbf{81.14} \\
        
        \bottomrule
    \end{tabular}
    
    \label{tab:scannetpp_shrinked}
\end{table*}

%% file: tables/resources_small_single_column.tex
\begin{table}
    \centering
    \caption{Time and memory complexity experiment on the Replica dataset, measured on a single NVIDIA H100 GPU at a fixed training budget of 200k iterations. MonoSDF, which does not support object-wise reconstruction, is included to illustrate the minimal overhead of our approach relative to multi-SDF approaches.}
    \large
    \setlength{\tabcolsep}{4pt}
    \begin{tabular}{@{}l c c c@{}}
        \toprule
        \textbf{Method} & \textbf{Time (h)} & \textbf{RAM (GB)} & \textbf{GPU (GB)} \\
        \midrule
        MonoSDF & 4.02 & 10.46 & 14.23 \\
        \midrule
        ObjectSDF++ & 9.88 & 13.59 & \textbf{9.33}  \\
        RICO & 10.97 & 11.69 & 25.87 \\
        FSTM & \textbf{4.27} & \textbf{10.66} & 14.66 \\        
        \bottomrule
    \end{tabular}

    
    \label{fig:resources}
\end{table}

%% file: tables/segmentation_2D.tex
\begin{table}
    \centering
    \caption{Colour renderings and 2D segmentation metrics for novel-view synthesis across the Replica dataset.}
    \large
    \begin{tabular}{@{}l c c@{}}
        \toprule
        \textbf{Method} & \textbf{PSNR}$\uparrow$ & \textbf{mIoU}$\uparrow$ \\
        \midrule

        ObjectSDF++ &23.87 & 0.836 \\
        RICO & 23.44 & 0.862 \\
        FSTM & \textbf{23.88} & \textbf{0.920} \\        
        \bottomrule
    \end{tabular}

    
    \label{tab:2d_seg}
\end{table}

%% file: main.bib
@String(CVPR= {IEEE Conf. Comput. Vis. Pattern Recog.})

@String(ICCV= {Int. Conf. Comput. Vis.})

@String(ECCV= {Eur. Conf. Comput. Vis.})

@String(ICLR = {Int. Conf. Learn. Represent.})

@String(IJCAI = {IJCAI})

@String(CVPR  = {CVPR})

@String(ICCV  = {ICCV})

@String(ECCV  = {ECCV})

@String(ICLR  = {ICLR})

@article{mildenhall_2021,
author = {Mildenhall, Ben and Srinivasan, Pratul P. and Tancik, Matthew and Barron, Jonathan T. and Ramamoorthi, Ravi and Ng, Ren},
title = {NeRF: representing scenes as neural radiance fields for view synthesis},
year = {2021},
issue_date = {January 2022},
publisher = {Association for Computing Machinery},
address = {New York, NY, USA},
volume = {65},
number = {1},
issn = {0001-0782},
journal = {Commun. ACM},
month = dec,
pages = {99–106},
numpages = {8}
}

@INPROCEEDINGS{panoptic_nerf,
  author={Fu, Xiao and Zhang, Shangzhan and Chen, Tianrun and Lu, Yichong and Zhu, Lanyun and Zhou, Xiaowei and Geiger, Andreas and Liao, Yiyi},
  booktitle={Int. Conf. 3D Vis. (3DV)}, 
  title={Panoptic NeRF: 3D-to-2D Label Transfer for Panoptic Urban Scene Segmentation}, 
  year={2022},
  volume={},
  number={},
  pages={1-11},
}

@InProceedings{Siddiqui_2023_CVPR,
    author    = {Siddiqui, Yawar and Porzi, Lorenzo and Bul\`o, Samuel Rota and M\"uller, Norman and Nie{\ss}ner, Matthias and Dai, Angela and Kontschieder, Peter},
    title     = {Panoptic Lifting for 3D Scene Understanding With Neural Fields},
    booktitle = {Proc. IEEE/CVF Conf. Comput. Vis. Pattern Recognit. (CVPR)},
    month     = {June},
    year      = {2023},
    pages     = {9043-9052}
}

@article{li2024monogsdf,
  title={MonoGSDF: Exploring Monocular Geometric Cues for Gaussian Splatting-Guided Implicit Surface Reconstruction},
  author={Li, Kunyi and Niemeyer, Michael and Chen, Zeyu and Navab, Nassir and Tombari, Federico},
  journal={arXiv preprint arXiv:2411.16898},
  year={2024}
}

@InProceedings{Liu_2023_CVPR,
    author    = {Liu, Fangfu and Zhang, Chubin and Zheng, Yu and Duan, Yueqi},
    title     = {Semantic Ray: Learning a Generalizable Semantic Field With Cross-Reprojection Attention},
    booktitle = {Proc. IEEE/CVF Conf. Comput. Vis. Pattern Recognit. (CVPR)},
    month     = {June},
    year      = {2023},
    pages     = {17386-17396}
}

@InProceedings{Zhi_2021_ICCV,
    author    = {Zhi, Shuaifeng and Laidlow, Tristan and Leutenegger, Stefan and Davison, Andrew J.},
    title     = {In-Place Scene Labelling and Understanding With Implicit Scene Representation},
    booktitle = {Proc. IEEE/CVF Int. Conf. Comput. Vis. (ICCV)},
    month     = {October},
    year      = {2021},
    pages     = {15838-15847}
}

@InProceedings{Oechsle_2021_ICCV,
    author    = {Oechsle, Michael and Peng, Songyou and Geiger, Andreas},
    title     = {UNISURF: Unifying Neural Implicit Surfaces and Radiance Fields for Multi-View Reconstruction},
    booktitle = {Proc. IEEE/CVF Int. Conf. Comput. Vis. (ICCV)},
    month     = {October},
    year      = {2021},
    pages     = {5589-5599}
}

@article{wang2021neus,
  title={Neus: Learning neural implicit surfaces by volume rendering for multi-view reconstruction},
  author={Wang, Peng and Liu, Lingjie and Liu, Yuan and Theobalt, Christian and Komura, Taku and Wang, Wenping},
  journal={arXiv preprint arXiv:2106.10689},
  year={2021}
}

@article{NEURIPS2020_1a77befc,
  title={Multiview neural surface reconstruction by disentangling geometry and appearance},
  author={Yariv, Lior and Kasten, Yoni and Moran, Dror and Galun, Meirav and Atzmon, Matan and Ronen, Basri and Lipman, Yaron},
  journal={Adv. Neural Inf. Process. Syst.},
  volume={33},
  pages={2492--2502},
  year={2020}
}

@inproceedings{NEURIPS2021_25e2a30f,
 author = {Yariv, Lior and Gu, Jiatao and Kasten, Yoni and Lipman, Yaron},
 title = {Volume Rendering of Neural Implicit Surfaces},
 booktitle = {Adv. Neural Inf. Process. Syst.},
 pages = {4805--4815},
 volume = {34},
 year = {2021}
}

@inproceedings{NEURIPS2022_16415eed,
 title = {Geo-Neus: Geometry-Consistent Neural Implicit Surfaces Learning for Multi-view Reconstruction},
 author = {Fu, Qiancheng and Xu, Qingshan and Ong, Yew Soon and Tao, Wenbing},
 booktitle = {Adv. Neural Inf. Process. Syst.},
 pages = {3403--3416},
 volume = {35},
 year = {2022}
}

@InProceedings{Darmon_2022_CVPR,
    author    = {Darmon, Fran\c{c}ois and Bascle, B\'en\'edicte and Devaux, Jean-Cl\'ement and Monasse, Pascal and Aubry, Mathieu},
    title     = {Improving Neural Implicit Surfaces Geometry With Patch Warping},
    booktitle = {Proc. IEEE/CVF Conf. Comput. Vis. Pattern Recognit. (CVPR)},
    month     = {June},
    year      = {2022},
    pages     = {6260-6269}
}

@InProceedings{Niemeyer_2022_CVPR,
    author    = {Niemeyer, Michael and Barron, Jonathan T. and Mildenhall, Ben and Sajjadi, Mehdi S. M. and Geiger, Andreas and Radwan, Noha},
    title     = {RegNeRF: Regularizing Neural Radiance Fields for View Synthesis From Sparse Inputs},
    booktitle = {Proc. IEEE/CVF Conf. Comput. Vis. Pattern Recognit. (CVPR)},
    month     = {June},
    year      = {2022},
    pages     = {5480-5490}
}

@InProceedings{Li_2023_CVPR,
    author    = {Li, Zhaoshuo and M\"uller, Thomas and Evans, Alex and Taylor, Russell H. and Unberath, Mathias and Liu, Ming-Yu and Lin, Chen-Hsuan},
    title     = {Neuralangelo: High-Fidelity Neural Surface Reconstruction},
    booktitle = {Proc. IEEE/CVF Conf. Comput. Vis. Pattern Recognit. (CVPR)},
    month     = {June},
    year      = {2023},
    pages     = {8456-8465}
}

@InProceedings{Guo_2022_CVPR,
    author    = {Guo, Haoyu and Peng, Sida and Lin, Haotong and Wang, Qianqian and Zhang, Guofeng and Bao, Hujun and Zhou, Xiaowei},
    title     = {Neural 3D Scene Reconstruction With the Manhattan-World Assumption},
    booktitle = {Proc. IEEE/CVF Conf. Comput. Vis. Pattern Recognit. (CVPR)},
    month     = {June},
    year      = {2022},
    pages     = {5511-5520}
}

@inproceedings{NEURIPS2022_9f0b1220,
 author = {Yu, Zehao and Peng, Songyou and Niemeyer, Michael and Sattler, Torsten and Geiger, Andreas},
 title = {MonoSDF: Exploring Monocular Geometric Cues for Neural Implicit Surface Reconstruction},
 booktitle = {Adv. Neural Inf. Process. Syst.},
 pages = {25018--25032},
 volume = {35},
 year = {2022}
}

@InProceedings{Wu_2023_ICCV,
    author    = {Wu, Qianyi and Wang, Kaisiyuan and Li, Kejie and Zheng, Jianmin and Cai, Jianfei},
    title     = {ObjectSDF++: Improved Object-Compositional Neural Implicit Surfaces},
    booktitle = {Proc. IEEE/CVF Int. Conf. Comput. Vis. (ICCV)},
    month     = {October},
    year      = {2023},
    pages     = {21764-21774}
}

@article{replica19arxiv,
  title={The {R}eplica Dataset: A Digital Replica of Indoor Spaces},
  author={Straub, Julian and Whelan, Thomas and Ma, Lingni and Chen, Yufan and Wijmans, Erik and Green, Simon and Engel, Jakob J and Mur-Artal, Raul and Ren, Carl and Verma, Shobhit and others},
  journal={arXiv preprint arXiv:1906.05797},
  year={2019}
}

@InProceedings{Park_2019_CVPR,
author = {Park, Jeong Joon and Florence, Peter and Straub, Julian and Newcombe, Richard and Lovegrove, Steven},
title = {DeepSDF: Learning Continuous Signed Distance Functions for Shape Representation},
booktitle = {Proc. IEEE/CVF Conf. Comput. Vis. Pattern Recognit. (CVPR)},
month = {June},
year = {2019}
}

@InProceedings{Zhang_2021_ICCV,
    author    = {Zhang, Jingyang and Yao, Yao and Quan, Long},
    title     = {Learning Signed Distance Field for Multi-View Surface Reconstruction},
    booktitle = {Proc. IEEE/CVF Int. Conf. Comput. Vis. (ICCV)},
    month     = {October},
    year      = {2021},
    pages     = {6525-6534}
}

@InProceedings{Zhang_2022_CVPR,
    author    = {Zhang, Jingyang and Yao, Yao and Li, Shiwei and Fang, Tian and McKinnon, David and Tsin, Yanghai and Quan, Long},
    title     = {Critical Regularizations for Neural Surface Reconstruction in the Wild},
    booktitle = {Proc. IEEE/CVF Conf. Comput. Vis. Pattern Recognit. (CVPR)},
    month     = {June},
    year      = {2022},
    pages     = {6270-6279}
}

@inproceedings{wang2022neuris,
  title={Neuris: Neural reconstruction of indoor scenes using normal priors},
  author={Wang, Jiepeng and Wang, Peng and Long, Xiaoxiao and Theobalt, Christian and Komura, Taku and Liu, Lingjie and Wang, Wenping},
  booktitle={Eur. Conf. Comput. Vis. (ECCV)},
  pages={139--155},
  year={2022},
  organization={Springer}
}

@InProceedings{Jain_2021_ICCV,
    author    = {Jain, Ajay and Tancik, Matthew and Abbeel, Pieter},
    title     = {Putting NeRF on a Diet: Semantically Consistent Few-Shot View Synthesis},
    booktitle = {Proc. IEEE/CVF Int. Conf. Comput. Vis. (ICCV)},
    month     = {October},
    year      = {2021},
    pages     = {5885-5894}
}

@inproceedings{xu2022sinnerf,
  title={Sinnerf: Training neural radiance fields on complex scenes from a single image},
  author={Xu, Dejia and Jiang, Yifan and Wang, Peihao and Fan, Zhiwen and Shi, Humphrey and Wang, Zhangyang},
  booktitle={Eur. Conf. Comput. Vis. (ECCV)},
  pages={736--753},
  year={2022},
  organization={Springer}
}

@inproceedings{wu2022object,
  title={Object-compositional neural implicit surfaces},
  author={Wu, Qianyi and Liu, Xian and Chen, Yuedong and Li, Kejie and Zheng, Chuanxia and Cai, Jianfei and Zheng, Jianmin},
  booktitle={Eur. Conf. Comput. Vis. (ECCV)},
  pages={197--213},
  year={2022},
  organization={Springer}
}

@InProceedings{Yang_2021_ICCV,
    author    = {Yang, Bangbang and Zhang, Yinda and Xu, Yinghao and Li, Yijin and Zhou, Han and Bao, Hujun and Zhang, Guofeng and Cui, Zhaopeng},
    title     = {Learning Object-Compositional Neural Radiance Field for Editable Scene Rendering},
    booktitle = {Proc. IEEE/CVF Int. Conf. Comput. Vis. (ICCV)},
    month     = {October},
    year      = {2021},
    pages     = {13779-13788}
}

@inproceedings{zha2023endosurf,
  title={Endosurf: Neural surface reconstruction of deformable tissues with stereo endoscope videos},
  author={Zha, Ruyi and Cheng, Xuelian and Li, Hongdong and Harandi, Mehrtash and Ge, Zongyuan},
  booktitle={Int. Conf. Med. Image Comput. Comput.-Assist. Interv. (MICCAI)},
  pages={13--23},
  year={2023},
  organization={Springer}
}

@InProceedings{Eftekhar_2021_ICCV,
    author    = {Eftekhar, Ainaz and Sax, Alexander and Malik, Jitendra and Zamir, Amir},
    title     = {Omnidata: A Scalable Pipeline for Making Multi-Task Mid-Level Vision Datasets From 3D Scans},
    booktitle = {Proc. IEEE/CVF Int. Conf. Comput. Vis. (ICCV)},
    month     = {October},
    year      = {2021},
    pages     = {10786-10796}
}

@misc{Yu2022SDFStudio,
    author    = {Yu, Zehao and Chen, Anpei and Antic, Bozidar and Peng, Songyou and Bhattacharyya, Apratim 
                 and Niemeyer, Michael and Tang, Siyu and Sattler, Torsten and Geiger, Andreas},
    title     = {SDFStudio: A Unified Framework for Surface Reconstruction},
    year      = {2022},
    url       = {https://github.com/autonomousvision/sdfstudio},
}

@inproceedings{yeshwanthliu2023scannetpp,
  title={ScanNet++: A High-Fidelity Dataset of 3D Indoor Scenes},
  author={Yeshwanth, Chandan and Liu, Yueh-Cheng and Nie{\ss}ner, Matthias and Dai, Angela},
  booktitle = {Proc. IEEE/CVF Int. Conf. Comput. Vis. ({ICCV})},
  pages={12--22},
  year={2023}
}

@article{kobayashi2022decomposing,
  title={Decomposing nerf for editing via feature field distillation},
  author={Kobayashi, Sosuke and Matsumoto, Eiichi and Sitzmann, Vincent},
  journal={Adv. Neural Inf. Process. Syst.},
  volume={35},
  pages={23311--23330},
  year={2022}
}

@ARTICLE{ov_nerf2024,
  author={Liao, Guibiao and Zhou, Kaichen and Bao, Zhenyu and Liu, Kanglin and Li, Qing},
  journal={IEEE Trans. Circuits Syst. Video Technol.}, 
  title={OV-NeRF: Open-Vocabulary Neural Radiance Fields With Vision and Language Foundation Models for 3D Semantic Understanding}, 
  year={2024},
  volume={34},
  number={12},
  pages={12923-12936},
}

@inproceedings{NEURIPS2023_525d2440,
 author = {Cen, Jiazhong and Zhou, Zanwei and Fang, Jiemin and yang, chen and Shen, Wei and Xie, Lingxi and Jiang, Dongsheng and Zhang, Xiaopeng and Tian, Qi},
 booktitle = {Adv. Neural Inf. Process. Syst.},
 editor = {A. Oh and T. Naumann and A. Globerson and K. Saenko and M. Hardt and S. Levine},
 pages = {25971--25990},
 publisher = {Curran Associates, Inc.},
 title = {Segment Anything in 3D with NeRFs},
 volume = {36},
 year = {2023}
}

@inproceedings{repaint_nerf,
author = {Zhou, Xingcheng and He, Ying and Yu, F. Richard and Li, Jianqiang and Li, You},
title = {RePaint-NeRF: NeRF editting via semantic masks and diffusion models},
year = {2023},
isbn = {978-1-956792-03-4},
booktitle = {Proc. Int. Joint Conf. Artif. Intell.},
articleno = {201},
numpages = {9},
location = {Macao, P.R.China},
series = {IJCAI '23}
}

@InProceedings{Haque_2023_ICCV,
    author    = {Haque, Ayaan and Tancik, Matthew and Efros, Alexei A. and Holynski, Aleksander and Kanazawa, Angjoo},
    title     = {Instruct-NeRF2NeRF: Editing 3D Scenes with Instructions},
    booktitle = {Proc. IEEE/CVF Int. Conf. Comput. Vis. (ICCV)},
    month     = {October},
    year      = {2023},
    pages     = {19740-19750}
}

@InProceedings{Xie_2023_ICCV,
    author    = {Xie, Baao and Li, Bohan and Zhang, Zequn and Dong, Junting and Jin, Xin and Yang, Jingyu and Zeng, Wenjun},
    title     = {NaviNeRF: NeRF-based 3D Representation Disentanglement by Latent Semantic Navigation},
    booktitle = {Proc. IEEE/CVF Int. Conf. Comput. Vis. (ICCV)},
    month     = {October},
    year      = {2023},
    pages     = {17992-18002}
}

@article{gropp2020implicit,
  title={Implicit geometric regularization for learning shapes},
  author={Gropp, Amos and Yariv, Lior and Haim, Niv and Atzmon, Matan and Lipman, Yaron},
  journal={arXiv preprint arXiv:2002.10099},
  year={2020}
}

@article{nesf,
  title={Nesf: Neural semantic fields for generalizable semantic segmentation of 3d scenes},
  author={Vora, Suhani and Radwan, Noha and Greff, Klaus and Meyer, Henning and Genova, Kyle and Sajjadi, Mehdi SM and Pot, Etienne and Tagliasacchi, Andrea and Duckworth, Daniel},
  journal={arXiv preprint arXiv:2111.13260},
  year={2021}
}

@article{pytorch,
  title={Pytorch: An imperative style, high-performance deep learning library},
  author={Paszke, Adam and Gross, Sam and Massa, Francisco and Lerer, Adam and Bradbury, James and Chanan, Gregory and Killeen, Trevor and Lin, Zeming and Gimelshein, Natalia and Antiga, Luca and others},
  journal={Adv. Neural Inf. Process. Syst.},
  volume={32},
  year={2019}
}

@inproceedings{kingmaAdam,
  author       = {Diederik P. Kingma and
                  Jimmy Ba},
  editor       = {Yoshua Bengio and
                  Yann LeCun},
  title        = {Adam: {A} Method for Stochastic Optimization},
  booktitle    = {Int. Conf. Learn. Represent. {(ICLR)},
                  San Diego, CA, USA, May 7-9, 2015, Conference Track Proceedings},
  year         = {2015}
}

@inproceedings{wu2024clusteringsdf,
  title={ClusteringSDF: Self-organized neural implicit surfaces for 3D decomposition},
  author={Wu, Tianhao and Zheng, Chuanxia and Wu, Qianyi and Cham, Tat-Jen},
  booktitle={Eur. Conf. Comput. Vis. (ECCV)},
  pages={255--272},
  year={2024},
  organization={Springer}
}

@InProceedings{chierchia_2025_wacv,
  author={Chierchia, Remi and Lebrat, Leo and Ahmedt-Aristizabal, David and Salvado, Olivier and Fookes, Clinton and Cruz, Rodrigo Santa},
  booktitle={IEEE/CVF Winter Conf. Appl. Comput. Vis. (WACV)}, 
  title={SALVE: A 3D Reconstruction Benchmark of Wounds from Consumer-Grade Videos}, 
  year={2025},
  volume={},
  number={},
  pages={4205-4214},
}

@inproceedings{rosinol2023nerf,
  title={Nerf-slam: Real-time dense monocular slam with neural radiance fields},
  author={Rosinol, Antoni and Leonard, John J and Carlone, Luca},
  booktitle={IEEE/RSJ Int. Conf. Intell. Robots Syst. (IROS)},
  pages={3437--3444},
  year={2023},
  organization={IEEE}
}

@article{adamkiewicz2022vision,
  title={Vision-only robot navigation in a neural radiance world},
  author={Adamkiewicz, Michal and Chen, Timothy and Caccavale, Adam and Gardner, Rachel and Culbertson, Preston and Bohg, Jeannette and Schwager, Mac},
  journal={IEEE Robot. Autom. Lett.},
  volume={7},
  number={2},
  pages={4606--4613},
  year={2022},
  publisher={IEEE}
}

@inproceedings{lin2023magic3d,
  title={Magic3d: High-resolution text-to-3d content creation},
  author={Lin, Chen-Hsuan and Gao, Jun and Tang, Luming and Takikawa, Towaki and Zeng, Xiaohui and Huang, Xun and Kreis, Karsten and Fidler, Sanja and Liu, Ming-Yu and Lin, Tsung-Yi},
  booktitle={Proc. IEEE/CVF Conf. Comput. Vis. Pattern Recognit. (CVPR)},
  pages={300--309},
  year={2023}
}

@inproceedings{liu2024surroundsdf,
  title={Surroundsdf: Implicit 3d scene understanding based on signed distance field},
  author={Liu, Lizhe and Wang, Bohua and Xie, Hongwei and Liu, Daqi and Liu, Li and Tian, Zhiqiang and Yang, Kuiyuan and Wang, Bing},
  booktitle={Proc. IEEE/CVF Conf. Comput. Vis. Pattern Recognit. (CVPR)},
  pages={21614--21623},
  year={2024}
}

@inproceedings{zhu2024sni,
  title={Sni-slam: Semantic neural implicit slam},
  author={Zhu, Siting and Wang, Guangming and Blum, Hermann and Liu, Jiuming and Song, Liang and Pollefeys, Marc and Wang, Hesheng},
  booktitle={Proc. IEEE/CVF Conf. Comput. Vis. Pattern Recognit. (CVPR)},
  pages={21167--21177},
  year={2024}
}

@article{xiao2025neural,
  title={Neural Radiance Fields for the Real World: A Survey},
  author={Xiao, Wenhui and Chierchia, Remi and Cruz, Rodrigo Santa and Li, Xuesong and Ahmedt-Aristizabal, David and Salvado, Olivier and Fookes, Clinton and Lebrat, Leo},
  journal={arXiv preprint arXiv:2501.13104},
  year={2025}
}

@InProceedings{Li_2023_ICCV,
    author    = {Li, Zizhang and Lyu, Xiaoyang and Ding, Yuanyuan and Wang, Mengmeng and Liao, Yiyi and Liu, Yong},
    title     = {RICO: Regularizing the Unobservable for Indoor Compositional Reconstruction},
    booktitle = {Proc. IEEE/CVF Int. Conf. Comput. Vis. (ICCV)},
    month     = {October},
    year      = {2023},
    pages     = {17761-17771}
}

@inproceedings{NEURIPS2024_2d880acd,
 author = {Ni, Junfeng and Chen, Yixin and Jing, Bohan and Jiang, Nan and Wang, Bin and Dai, Bo and Li, Puhao and Zhu, Yixin and Zhu, Song-Chun and Huang, Siyuan},
 title = {PhyRecon: Physically Plausible Neural Scene Reconstruction},
 booktitle = {Adv. Neural Inf. Process. Syst.},
 pages = {25747--25780},
 publisher = {Curran Associates, Inc.},
 volume = {37},
 year = {2024}
}

@InProceedings{Lyu_2024_CVPR,
    author    = {Lyu, Xiaoyang and Chang, Chirui and Dai, Peng and Sun, Yang-Tian and Qi, Xiaojuan},
    title     = {Total-Decom: Decomposed 3D Scene Reconstruction with Minimal Interaction},
    booktitle = {Proc. IEEE/CVF Conf. Comput. Vis. Pattern Recognit. (CVPR)},
    month     = {June},
    year      = {2024},
    pages     = {20860-20869}
}

@article{park2024h2o,
  title={H2O-SDF: two-phase learning for 3D indoor reconstruction using object surface fields},
  author={Park, Minyoung and Do, Mirae and Shin, YeonJae and Yoo, Jaeseok and Hong, Jongkwang and Kim, Joongrock and Lee, Chul},
  journal={arXiv preprint arXiv:2402.08138},
  year={2024}
}

@InProceedings{Ni_2025_CVPR,
    author    = {Ni, Junfeng and Liu, Yu and Lu, Ruijie and Zhou, Zirui and Zhu, Song-Chun and Chen, Yixin and Huang, Siyuan},
    title     = {Decompositional Neural Scene Reconstruction with Generative Diffusion Prior},
    booktitle = {Proc. IEEE/CVF Conf. Comput. Vis. Pattern Recognit. (CVPR)},
    month     = {June},
    year      = {2025},
    pages     = {6022-6033}
}

@article{ren2025fin3r,
  title={Fin3R: Fine-tuning Feed-forward 3D Reconstruction Models via Monocular Knowledge Distillation},
  author={Ren, Weining and Wang, Hongjun and Tan, Xiao and Han, Kai},
  journal={arXiv preprint arXiv:2511.22429},
  year={2025}
}

@inproceedings{tang2025mv,
  title={Mv-dust3r+: Single-stage scene reconstruction from sparse views in 2 seconds},
  author={Tang, Zhenggang and Fan, Yuchen and Wang, Dilin and Xu, Hongyu and Ranjan, Rakesh and Schwing, Alexander and Yan, Zhicheng},
  booktitle={Proc. IEEE/CVF Conf. Comput. Vis. Pattern Recognit. (CVPR)},
  pages={5283--5293},
  year={2025}
}

@article{wang2025plgs,
  title={Plgs: Robust panoptic lifting with 3d gaussian splatting},
  author={Wang, Yu and Wei, Xiaobao and Lu, Ming and Kang, Guoliang},
  journal={IEEE Trans. Image Process.},
  year={2025},
  volume={34},
  pages={3377-3388},
  publisher={IEEE}
}

@inproceedings{xie2025panopticsplatting,
  title={Panopticsplatting: End-to-end panoptic gaussian splatting},
  author={Xie, Yuxuan and Yu, Xuan and Jiang, Changjian and Mao, Sitong and Zhou, Shunbo and Fan, Rui and Xiong, Rong and Wang, Yue},
  booktitle={IEEE/RSJ Int. Conf. Intell. Robots Syst. (IROS)},
  pages={4067--4074},
  year={2025},
  organization={IEEE}
}

@inproceedings{zhu2025objectgs,
  title={Objectgs: Object-aware scene reconstruction and scene understanding via gaussian splatting},
  author={Zhu, Ruijie and Yu, Mulin and Xu, Linning and Jiang, Lihan and Li, Yixuan and Zhang, Tianzhu and Pang, Jiangmiao and Dai, Bo},
  booktitle={Proc. IEEE/CVF Int. Conf. Comput. Vis. (ICCV)},
  pages={8350--8360},
  year={2025}
}

@inproceedings{ye2024gaussian,
  title={Gaussian grouping: Segment and edit anything in 3d scenes},
  author={Ye, Mingqiao and Danelljan, Martin and Yu, Fisher and Ke, Lei},
  booktitle={Eur. Conf. Comput. Vis. (ECCV)},
  pages={162--179},
  year={2024},
  organization={Springer}
}

@article{kerbl20233d,
  title={3d gaussian splatting for real-time radiance field rendering.},
  author={Kerbl, Bernhard and Kopanas, Georgios and Leimk{\"u}hler, Thomas and Drettakis, George and others},
  journal={ACM Trans. Graph.},
  volume={42},
  number={4},
  pages={139--1},
  year={2023}
}

@article{Sitzmann2020ImplicitFunctions,
    title = {{Implicit Neural Representations with Periodic Activation Functions}},
    year = {2020},
    author = {Sitzmann, Vincent and Martel, Julien N. P. and Bergman, Alexander W. and Lindell, David B. and Wetzstein, Gordon},
    month = {6},
    url = {http://arxiv.org/abs/2006.09661},
    arxivId = {2006.09661}
}

@article{attene2010lightweight,
  title={A lightweight approach to repairing digitized polygon meshes},
  author={Attene, Marco},
  journal={The visual computer},
  volume={26},
  number={11},
  pages={1393--1406},
  year={2010},
  publisher={Springer}
}
